\documentclass[10pt,journal,compsoc]{IEEEtran}
\usepackage{cite}

\ifCLASSINFOpdf
  \usepackage[pdftex]{graphicx}
  \graphicspath{{./images/}}
  \DeclareGraphicsExtensions{.pdf,.jpeg,.jpg,.png}
\else
\fi
\usepackage{amsmath}
\usepackage{amsthm}
\usepackage{amsfonts}

\newcommand{\X}{X}
\newcommand{\y}{y}
\newcommand{\T}{t}

\newcommand{\Prob}{P}
\newcommand{\Sample}{S}
\newcommand{\Distribution}{F}
\newcommand{\Data}{d}

\ifCLASSOPTIONcompsoc
 \usepackage[caption=false,font=normalsize,labelfont=sf,textfont=sf]{subfig}
\else
 \usepackage[caption=false,font=footnotesize]{subfig}
\fi
\usepackage{url}

\usepackage{color}
\usepackage{soul}

\newcommand{\hly}[2][yellow]{{\sethlcolor{white}\hl{#2}}}

\soulregister\cite7
\soulregister\ref7
\soulregister\pageref7
\sethlcolor{white}
\newcommand\MYhyperrefoptionsA{bookmarks=true,bookmarksnumbered=true,
pdfpagemode={UseOutlines},plainpages=false,pdfpagelabels=true,
colorlinks=true,linkcolor={black},citecolor={black},urlcolor={black},
pdftitle={Bare Demo of IEEEtran.cls for Computer Society Journals},
pdfsubject={Typesetting},
pdfauthor={Michael D. Shell},
pdfkeywords={Computer Society, IEEEtran, journal, LaTeX, paper,
             template}}

\ifCLASSINFOpdf
  \usepackage[\MYhyperrefoptionsA,pdftex]{hyperref}
\else
  \usepackage[\MYhyperrefoptionsA,breaklinks=true,dvips]{hyperref}
  \usepackage{breakurl}
\fi

\usepackage{multirow}
\usepackage{multicol}

\usepackage{color}
\usepackage{soul}

\begin{document}
%
\title{Learning under Concept Drift: A Review}
%
%
%

\author{Jie~Lu,~\IEEEmembership{Fellow,~IEEE,}
        Anjin~Liu,~\IEEEmembership{Member,~IEEE,}
        Fan~Dong,
        Feng~Gu,
        Jo\~{a}o~Gama,
        and~Guangquan~Zhang
}

\IEEEtitleabstractindextext{%
\begin{abstract}
Concept drift describes unforeseeable changes in the underlying distribution of streaming data over time. Concept drift research involves the development of methodologies and techniques for drift detection, understanding and adaptation. Data analysis has revealed that machine learning in a concept drift environment will result in poor learning results if the drift is not addressed. To help researchers identify which research topics are significant and how to apply related techniques in data analysis tasks, it is necessary that a high quality, instructive review of current research developments and trends in the concept drift field is conducted. In addition, due to the rapid development of concept drift in recent years, the methodologies of learning under concept drift have become noticeably systematic, unveiling a framework which has not been mentioned in literature. This paper reviews over 130 high quality publications in concept drift related research areas, analyzes up-to-date developments in methodologies and techniques, and establishes a framework of learning under concept drift including three main components: concept drift detection, concept drift understanding, and concept drift adaptation. This paper lists and discusses 10 popular synthetic datasets and 14 publicly available benchmark datasets used for evaluating the performance of learning algorithms aiming at handling concept drift. Also, concept drift related research directions are covered and discussed. By providing state-of-the-art knowledge, this survey will directly support researchers in their understanding of research developments in the field of learning under concept drift.
\end{abstract}

}

\maketitle





\IEEEdisplaynontitleabstractindextext

%
\IEEEpeerreviewmaketitle

%
%
%
%







\IEEEraisesectionheading{\section{Introduction}
\label{sce-in}}

\IEEEPARstart{G}{overnments} and companies are generating huge amounts of streaming data and urgently need efficient data analytics and machine learning techniques to support them making predictions and decisions. However, the rapidly changing environment of new products, new markets and new customer behaviors inevitably results in the appearance of concept drift problem. Concept drift means that the statistical properties of the target variable, which the model is trying to predict, change over time in unforeseen ways \cite{W:machinelearning}. If the concept drift occurs, the induced pattern of past data may not be relevant to the new data, leading to poor predictions and decision outcomes. The phenomenon of concept drift has been recognized as the root cause of decreased effectiveness in many data-driven information systems such as data-driven early warning systems and data-driven decision support systems. In an ever-changing and big data environment, how to provide more reliable data-driven predictions and decision facilities has become a crucial issue.

\begin{figure}[!th]
\centering
\includegraphics[width=0.48\textwidth]{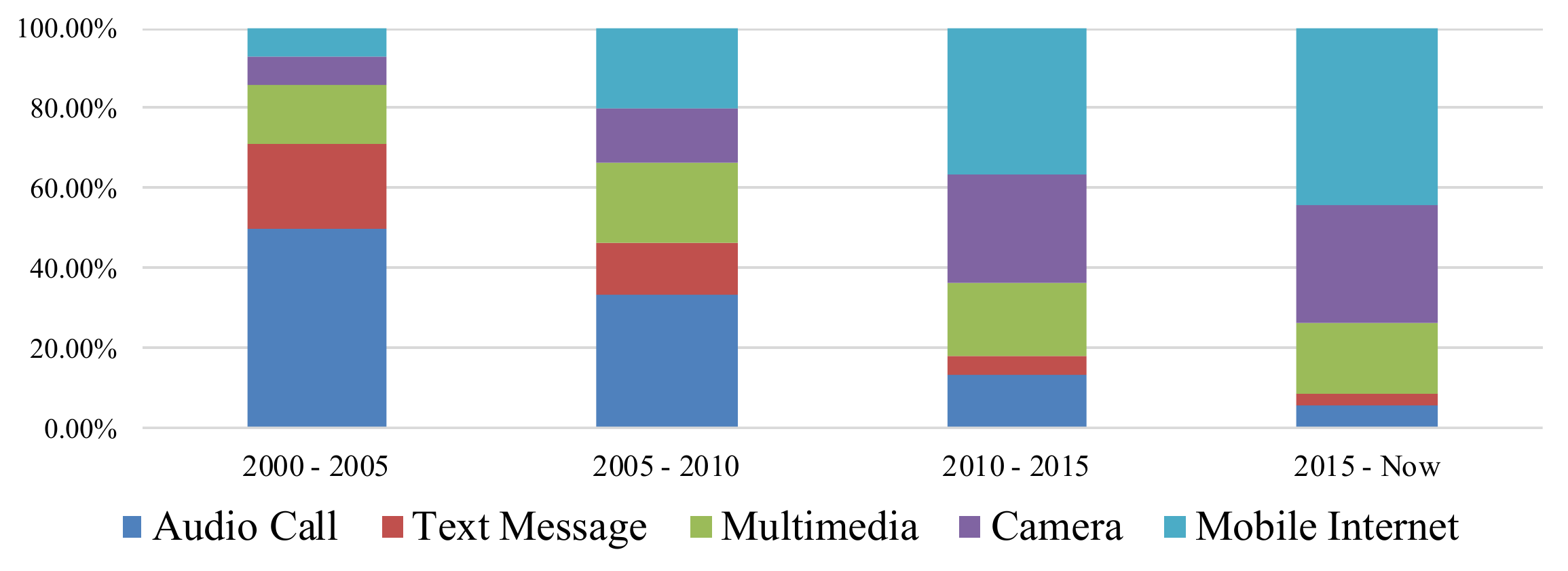}
\caption{Concept drift in mobile phone usage (data used in figure are for demonstration only)}
\label{fig-mobile-phone-demo}
\end{figure}

\hly{Concept drift problem exists in many real-world situations. An example can be seen in the changes of behavior in mobile phone usage, as shown in \figurename~\ref{fig-mobile-phone-demo}.
From the bars in this figure, the time percentage distribution of the mobile phone usage pattern has changed from ``Audio Call'' to ``Camera'' and then to ``Mobile Internet'' over the past two decades.}


Recent attractive research in the field of concept drift targets more challenging problems, i.e., how to accurately detect concept drift in unstructured and noisy datasets \cite{L:AI2, L:competence}, how to quantitatively understand concept drift in a explainable way \cite{Liu:IJCAI2017, Liu:FuzzyWindow}, and how to effectively react to drift by adapting related knowledge \cite{krawczyk2017ensemble, ramirez2017a}.

Solving these challenges endows prediction and decision-making with the adaptability in an uncertain environment. Conventional research related to machine learning has been significantly improved by introducing concept drift techniques in data science and artificial intelligence in general, and in pattern recognition and data stream mining in particular. These new studies enhance the effectiveness of analogical and knowledge reasoning in an ever-changing environment. A new topic is formed during this development: adaptive data-driven prediction/decision systems. In particular, concept drift is a highly prominent and significant issue in the context of the big data era because the uncertainty of data types and data distribution is an inherent nature of big data.

\begin{figure}[!t]
\centering
\includegraphics[width=0.48\textwidth]{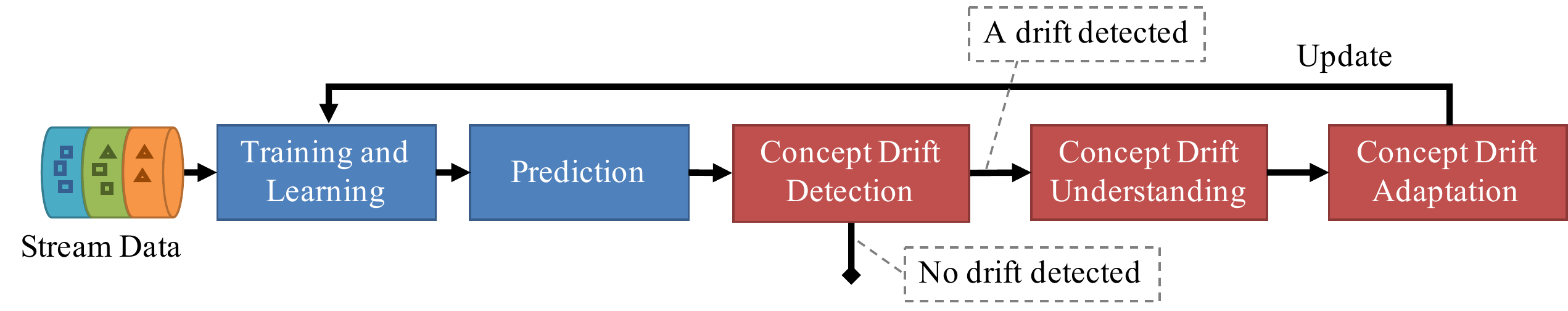}
\caption{Framework for handling concept drift in machine learning. 
\hly{Please note that some methods can do concept drift detection and concept drift understanding simultaneously.}
}
\label{fig-whole-framework}
\end{figure}

Conventional machine learning has two main components: training/learning and prediction. Research on learning under concept drift presents three new components: drift detection (whether or not drift occurs), drift understanding (when, how, where it occurs) and drift adaptation (reaction to the existence of drift) as shown in \figurename~\ref{fig-whole-framework}.
These will be discussed in Section~\ref{sec-dd}-\ref{sec-ka}.

In literature, a detailed concept drift survey paper \cite{gama2014survey} was published in 2014 but intentionally left certain sub-problems of concept drift to other publications, such as \hly{the details of the data distribution change ($\Prob(\X)$)} as mentioned in their Section 2.1. In 2015, another comprehensive survey paper \cite{Polikar:Survey} was published, which surveys and gives tutorial of both the established and the state-of-the-art approaches. It provides a hybrid-view about concept drift from two primary perspectives, active and passive. Both survey papers are comprehensive and can be a good introduction to concept drift researching. \hly{However, many new publications have become available in the last three years, even a new category of drift detection methods has arisen, named multiple hypothesis tests drift detection.} It is necessary to review the past research focuses and give the most recent research trends about concept drift, which is one of the main contribution of this survey paper.

Besides these two publications, four related survey papers \cite{gama2012survey, krawczyk2017ensemble, ramirez2017a, silva2013data} have also provided valuable insights into how to address concept drift, but their specific research focus is only on data stream learning, rather than analyzing concept drift adaptation algorithms and understanding concept drift. Specifically, paper \cite{ramirez2017a} focuses on data reduction for stream learning incorporating concept drift, while \cite{krawczyk2017ensemble} only focuses on investigating the development in learning ensembles for data stream learning in a dynamic environment. \cite{silva2013data} concerns the evolution of data stream clustering, and \cite{gama2012survey} focuses on investigating the current and future trends of data stream learning. There is therefore a gap in the current literature that requires a fuller picture of established and the new emerged research on concept drift; a comprehensive review of the three major aspects of concept drift: concept drift detection, understanding and adaptation, as shown in \figurename~\ref{fig-whole-framework}; and a discussion about the new trend of concept drift research.

The selection of references in this survey paper was performed according to the following steps:

Step 1. Publication database: Science Direct, ACM Digital Library, IEEE Xplore and SpringerLink.

Step 2. Preliminary screening of articles: The first search was based on keywords. The articles were then selected as references if they 1) present new theory, algorithm or methodology in the area of concept drift, or 2) report a concept drift application.

Step 3. Result filtering for articles: The articles selected in Step 2 were divided into three groups: concept drift detection, understanding, and adaptation. The references in each group were filtered again, based on 1) Time: published mainly within the last 10 years, or 2) Impact: published in high quality journals/conferences or having high citations.

Step 4. Dataset selection: To help readers test their research results, this paper lists popular datasets and their characteristics, the dataset providers, and how each dataset can be used.

On completion of this process, 137 research articles, 10 widely used synthetic datasets for evaluating the performance of learning algorithms dealing with concept drift, and 14 publicly available and widely used real-world datasets were listed for discussion.

The main contributions of this paper are:

\begin{enumerate}[\IEEEsetlabelwidth{6)}]

\item It perceptively summarizes concept drift research achievements and clusters the research into three categories: concept drift detection, understanding and adaptation, providing a clear framework for concept drift research development (\figurename~\ref{fig-whole-framework});


\item It proposes a new component, concept drift understanding, for retrieving information about the status of concept drift in aspects of when, how, and where. This also creates a connection between drift detection and drift adaptation;

\item It uncovers several very new concept drift techniques, such as active learning under concept drift and fuzzy competence model-based drift detection, and identifies related research involving concept drift;

\item It systematically examines two sets of concept drift datasets, Synthetic datasets and Real-world datasets, through multiple dimensions: dataset description, availability, suitability for type of drift, and existing applications;

\item It suggests several emerging research topics and potential research directions in this area.

\end{enumerate}

The remainder of this paper is structured as follows. In Section~\ref{sec-pd}, the definitions of concept drift are given and discussed. Section~\ref{sec-dd} presents research methods and algorithms in concept drift detection.
Section~\ref{sec-du} discusses research developments in concept drift understanding. Research results on drift adaptation (concept drift reaction) are reported in Section~\ref{sec-ka}.
Section~\ref{sec-eb} presents evaluation systems and related datasets used to test concept drift algorithms.
Section~\ref{sec-ra} summaries related research concerning the concept drift problem.
Section~\ref{sec-ca} presents a comprehensive analysis of main findings and future research directions. 

\section{Problem Description}
\label{sec-pd}

This section first gives the formal definition and the sources of concept drift in Section~\ref{sub-pd-definition-sources}. Then, in Section~\ref{sub-pd-types-dirft}, the commonly defined types of concept drift are introduced.

\subsection{Concept drift definition and the sources}
\label{sub-pd-definition-sources}

Concept drift is a phenomenon in which the statistical properties of a target domain change over time in an arbitrary way \cite{L:competence}. It was first proposed by \cite{schlimmer1986incremental} who aimed to point out that noise data may turn to non-noise information at different time. These changes might be caused by changes in hidden variables which cannot be measured directly \cite{Liu:IJCAI2017}. Formally, concept drift is defined as follows:

Given a time period $[0,\T]$, a set of samples, denoted as $\Sample_{0,\T}=\{\Data_{0},\dots ,\Data_{\T}\}$, where $\Data_{i}=(\X_i,\y_i)$ is one observation (or a data instance), $\X_i$ is the feature vector, $\y_i$ is the label, and $\Sample_{0,\T}$ follows a certain distribution $\Distribution_{0,\T}(\X,\y)$.
\hly{Concept drift occurs at timestamp $\T+1$, if $\Distribution_{0,\T}(\X,\y) \neq \Distribution_{\T+1,\infty}(\X,\y)$, denoted as $\exists \T\colon \Prob_{\T}(\X,\y) \neq \Prob_{\T+1}(\X,\y)$} \cite{gama2014survey, Viktor:SAMkNN, L:AI2, zliobaite2015}.

Concept drift has also been defined by various authors using alternative names, such as dataset shift \cite{amos2008when} or concept shift \cite{W:machinelearning}. Other related terminologies were introduced in \cite{Jose:ConceptShift}'s work, the authors proposed that concept drift or shift is only one subcategory of dataset shift and the dataset shift is consists of covariate shift, prior probability shift and concept shift. These definitions clearly stated the research scope of each research topics. However, since concept drift is usually associated with covariate shift and prior probability shift, and an increasing number of publications \cite{gama2014survey, Viktor:SAMkNN, L:AI2, zliobaite2015} refer to the term "concept drift" as the problem in which $\exists \T\colon \Prob_{\T}(\X,\y) \neq \Prob_{\T+1}(\X,\y)$. Therefore, we apply the same definition of concept drift in this survey. Accordingly, concept drift at time $\T$ can be defined as the change of joint probability of $\X$ and $\y$ at time $\T$. Since the joint probability $P_t(X,y)$ can be decomposed into two parts as $P_t(X,y)=P_t(X)\times P_t(y|X)$, concept drift can be triggered by three sources:

\begin{itemize}

\item \textbf{Source I}: $\Prob_{\T}(\X) \neq \Prob_{\T+1}(\X)$ while $\Prob_{\T}(\y|\X) = \Prob_{\T+1}(\y|\X)$, that is, the research focus is the drift in $\Prob_{\T}(\X)$ while $\Prob_{\T}(\y|\X)$ remains unchanged. Since $\Prob_{\T}(\X)$ drift does not affect the decision boundary, it has also been considered as virtual drift \cite{ramirez2017a}, Fig. 3(a).

\item \textbf{Source II}: $\Prob_{\T}(\y|\X) \neq \Prob_{\T+1}(\y|\X)$ while $\Prob_{\T}(\X) =\Prob_{\T+1}(\X)$ while $\Prob_{\T}(\X)$ remains unchanged. This drift will cause decision boundary change and lead to learning accuracy decreasing, which is also called actual drift, Fig. 3(b).

\item \textbf{Source III}: mixture of Source I and Source II, namely $\Prob_{\T}(\X) \neq \Prob_{\T+1}(\X)$ and $\Prob_{\T}(\y|\X) \neq \Prob_{\T+1}(\y|\X)$. Concept drift focus on the drift of both $\Prob_{\T}(\y|\X)$ and $\Prob_{\T}(\X)$, since both changes convey important information about learning environment Fig. 3(c).

\end{itemize}

\begin{figure}[!t]
\centering
\includegraphics[width=0.48\textwidth]{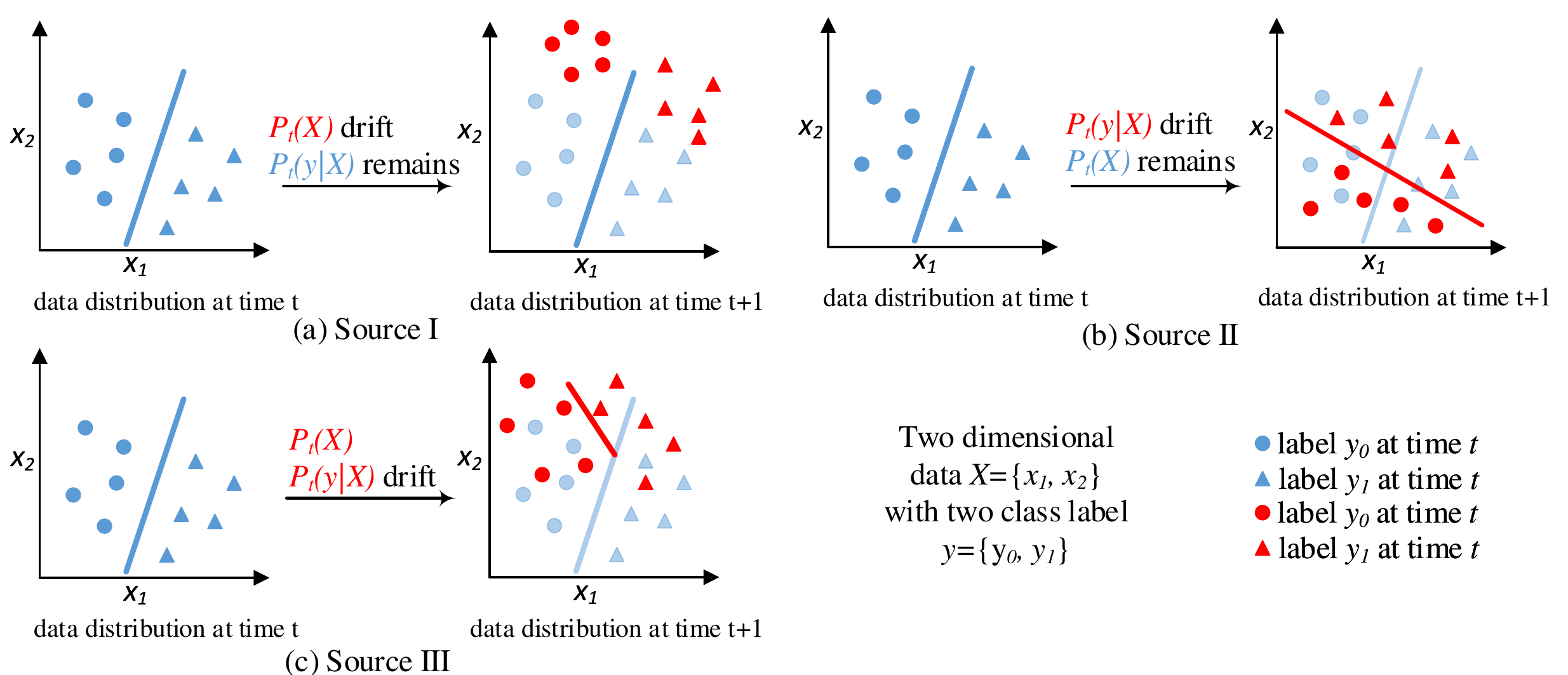}
\caption{Three sources of concept drift}
\label{fig-concept-drift-sources}
\end{figure}

\figurename~\ref{fig-concept-drift-sources} demonstrates how these sources differ from each other in a two-dimensional feature space. Source I is feature space drift, and Source II is decision boundary drift. In many real-world applications, Source I and Source II occur together, which creates Source III.

\subsection{The types of concept drift}
\label{sub-pd-types-dirft}

Commonly, concept drift can be distinguished as four types \cite{gama2014survey} as shown in \figurename~\ref{fig-concept-drift-types}:

 \begin{figure}[!t]
\centering
\includegraphics[width=0.45\textwidth]{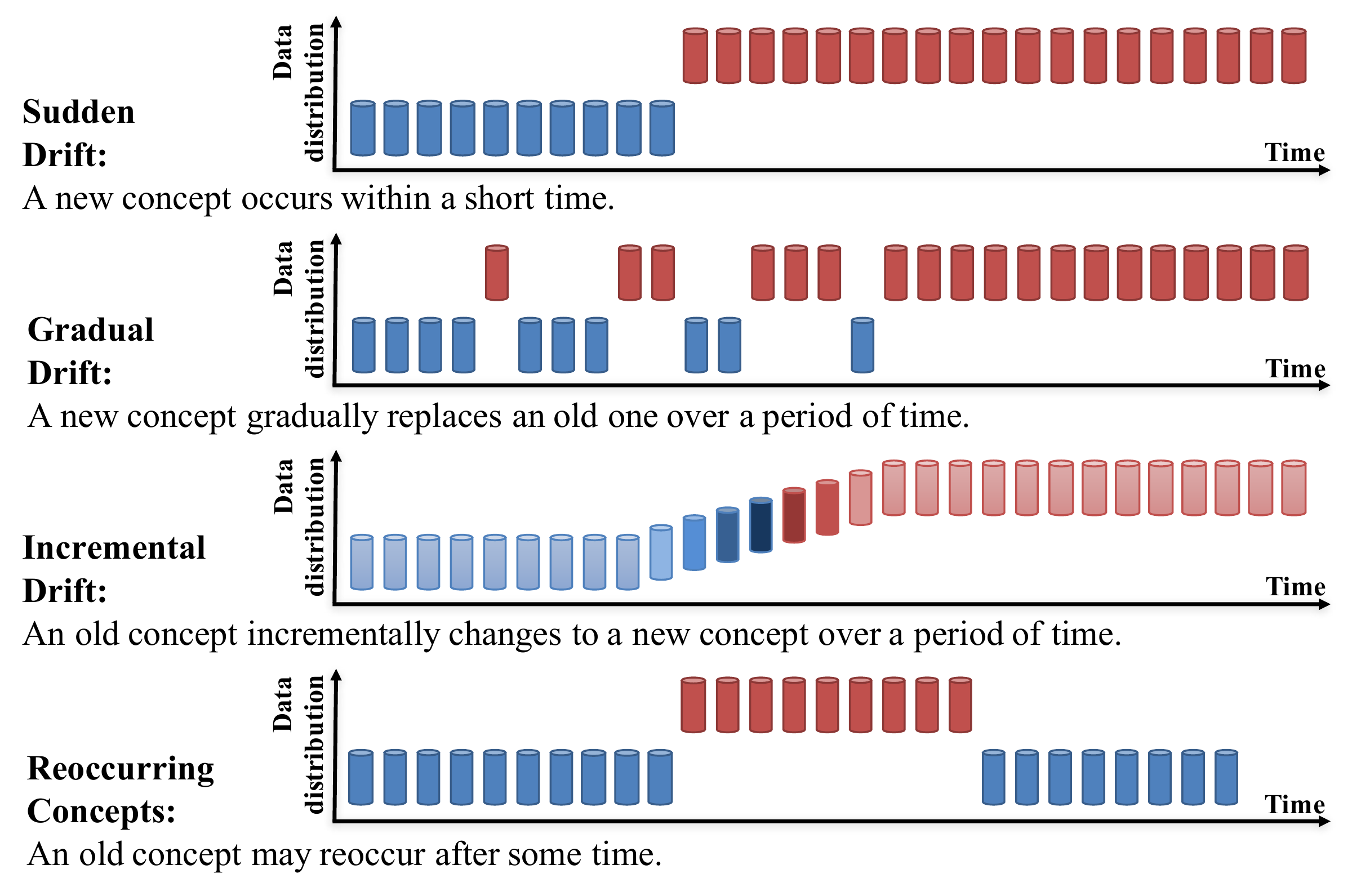}
\caption{An example of concept drift types.}
\label{fig-concept-drift-types}
\end{figure}

Research into concept drift adaptation in Types 1-3 focuses on how to minimize the drop in accuracy and achieve the fastest recovery rate during the concept transformation process. In contrast, the study of Type 4 drift emphasizes the use of historical concepts, that is, how to find the best matched historical concepts with the shortest time. The new concept may suddenly, incrementally, or gradually reoccur.

To better demonstrate the differences between these types, the term ``intermediate concept'' was introduced by \cite{gama2014survey} to describe the transformation between concepts. As mentioned by \cite{Liu:IJCAI2017}, a concept drift may not only take place at an exact timestamp, but may also last for a long period. As a result, intermediate concepts may appear during the transformation as one concept (starting concept) changes to another (ending concept). An intermediate concept can be a mixture of the starting concept and the ending concept, like the incremental drift, or one of the starting or ending concept, such as the gradual drift.


\section{Concept Drift detection}
\label{sec-dd}

This section focuses on summarizing concept drift detection algorithms. Section~\ref{sub-dd-framework} introduces a typical drift detection framework. Then, Section~\ref{sub-dd-algorithms} systematically reviews and categorizes drift detection algorithms according to their implementation details for each component in the framework. At last, Section \ref{sub-dd-summary} lists the state-of-the-art drift detection algorithms with comparisons of their implementation details.

\subsection{A general framework for drift detection}
\label{sub-dd-framework}

\begin{figure}[!t]
\centering
\includegraphics[width=0.44\textwidth]{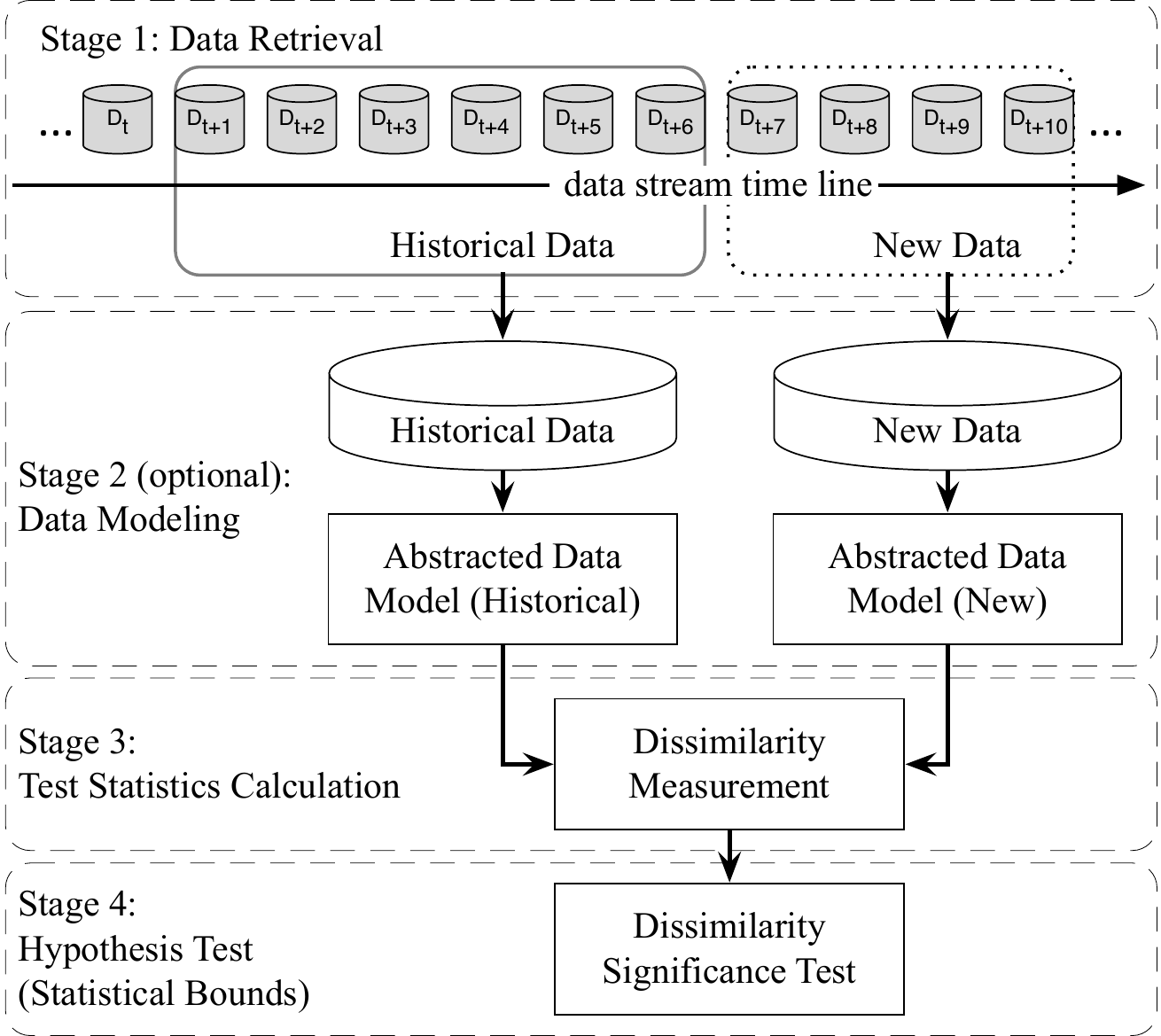}
\caption{An overall framework for concept drift detection}
\label{fig-framework-drift-detection}
\end{figure}

Drift detection refers to the techniques and mechanisms that characterize and quantify concept drift via identifying change points or change time intervals \cite{basseville1993detection}. A general framework for drift detection contains four stages, as shown in \figurename~\ref{fig-framework-drift-detection}.

Stage 1 (Data Retrieval) aims to retrieve data chunks from data streams. Since a single data instance cannot carry enough information to infer the overall distribution \cite{L:AI2}, knowing how to organize data chunks to form a meaningful pattern or knowledge is important in data stream analysis tasks \cite{ramirez2017a}.

Stage 2 (Data Modeling) aims to abstract the retrieved data and extract the key features containing sensitive information, that is, the features of the data that most impact a system if they drift. This stage is optional, because it mainly concerns dimensionality reduction, or sample size reduction, to meet storage and online speed requirements \cite{Liu:IJCAI2017}.

Stage 3 (Test Statistics Calculation) is the measurement of dissimilarity, or distance estimation. It quantifies the severity of the drift and forms test statistics for the hypothesis test. It is considered to be the most challenging aspect of concept drift detection. The problem of how to define an accurate and robust dissimilarity measurement is still an open question. A dissimilarity measurement can also be used in clustering evaluation \cite{silva2013data}, and to determine the dissimilarity between sample sets \cite{Dries:MMDconceptDriftDetection}.

Stage 4 (Hypothesis Test) uses a specific hypothesis test to evaluate the statistical significance of the change observed in Stage 3, or the p-value. They are used to determine drift detection accuracy by proving the statistical bounds of the test statistics proposed in Stage 3. Without Stage 4, the test statistics acquired in Stage 3 are meaningless for drift detection, because they cannot determine the drift confidence interval, that is, how likely it is that the change is caused by concept drift and not noise or random sample selection bias \cite{L:competence}. The most commonly used hypothesis tests are: estimating the distribution of the test statistics \cite{Alippi:JITpart1, Gama:DDM}, bootstrapping \cite{Alippi:LSDD-CDT, D:kdqTree}, the permutation test \cite{L:competence}, and Hoeffding's inequality-based bound identification \cite{I:HDDM}.

It is also worth to mention that, without Stage 1, the concept drift detection problem can be considered as a two-sample test problem which examines whether the population of two given sample sets are from the same distribution \cite{Dries:MMDconceptDriftDetection}. In other words, any multivariate two-sample test is an option that can be adopted in Stages 2-4 to detect concept drift \cite{Dries:MMDconceptDriftDetection}. However, in some cases, the distribution drift may not be included in the target features, therefore the selection of the target feature will affect the overall performance of a learning system and is a critical problem in concept drift detection \cite{Y:ChangePointFeatureSelection}.

\subsection{Concept drift detection algorithms}
\label{sub-dd-algorithms}

This section surveys drift detection methods and algorithms, which are classified into three categories in terms of the test statistics they apply.

\subsubsection{Error rate-based drift detection}
\label{sss-error-rate}

PLearner error rate-based drift detection algorithms form the largest category of algorithms. These algorithms focus on tracking changes in the online error rate of base classifiers. If an increase or decrease of the error rate is proven to be statistically significant, an upgrade process (drift alarm) will be triggered.

\begin{figure}[!t]
\centering
\includegraphics[width=0.48\textwidth]{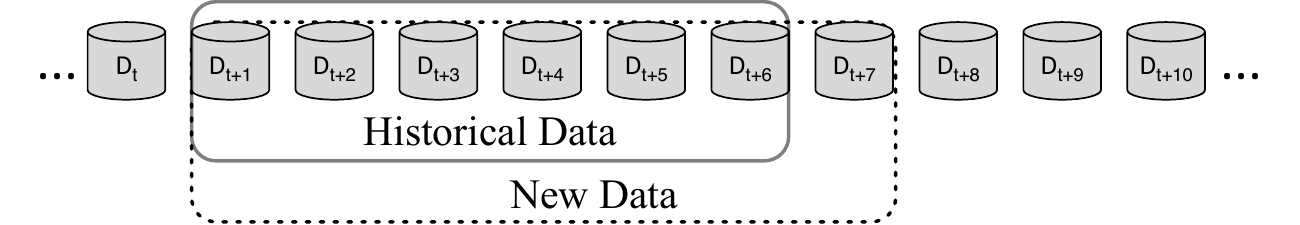}
\caption{Landmark time window for drift detection. The starting point of the window is fixed, while the end point of the window will be extended after a new data instance has been received.}
\label{fig-landmark-time-window}
\end{figure}

One of the most-referenced concept drift detection algorithms is the Drift Detection Method (DDM) \cite{Gama:DDM}. It was the first algorithm to define the warning level and drift level for concept drift detection. In this algorithm, Stage 1 is implemented by a landmark time window, as shown in \figurename~\ref{fig-landmark-time-window}. When a new data instance become available for evaluation, DDM detects whether the overall online error rate within the time window has increased significantly. If the confidence level of the observed error rate change reaches the warning level, DDM starts to build a new learner while using the old learner for predictions. If the change reached the drift level, the old learner will be replaced by the new learner for further prediction tasks. To acquire the online error rate, DDM needs a classifier to make the predictions. This process converts training data to a learning model, which is considered as the Stage 2 (Data Modeling). The test statistics in Stage 3 constitute the online error rate. The hypothesis test, Stage 4, is conducted by estimating the distribution of the online error rate and calculating the warning level and drift threshold.

Similar implementations have been adopted and applied in the Learning with Local Drift Detection (LLDD) \cite{Gama:local}, Early Drift Detection Method (EDDM) \cite{B:EDDM}, Heoffding's inequality based Drift Detection Method (HDDM) \cite{I:HDDM}, Fuzzy Windowing Drift Detection Method (FW-DDM) \cite{Liu:FuzzyWindow}, Dynamic Extreme Learning Machine (DELM) \cite{Xu:DELM}. LLDD modifies Stages 3 and 4, dividing the overall drift detection problem into a set of decision tree node-based drift detection problems; EDDM improves Stage 3 of DDM using the distance between two correct classifications to improve the sensitivity of drift detection; HDDM modifies Stage 4 using Hoeffding's inequality to identify the critical region of a drift; FW-DDM improves Stage 1 of DDM using a fuzzy time window instead of a conventional time window to address the gradual drift problem; DEML does not change the DDM detection algorithm but uses a novel base learner, which is a single hidden layer feedback neural network called Extreme Learning Machine (ELM)\cite{huang2006extreme} to improve the adaptation process after a drift has been confirmed. EWMA for Concept Drift Detection (ECDD) \cite{G:ECDD} takes advantage of the error rate to detect concept drift. ECDD employs the EWMA chart to track changes in the error rate. The implementation of Stages 1-3 of ECDD is the same as for DDM, while Stage 4 is different. ECDD modifies the conventional EWMA chart using a dynamic mean $\hat{p}_{0,t}$ instead of the conventional static mean $p_0$, where $\hat{p}_{0,t}$ is the estimated online error rate within time $[0,t]$, and $p_0$ implies the theoretical error rate when the learner was initially built. Accordingly, the dynamic variance can be calculated by $\sigma_{Z_{t}}^{2} = \hat{p}_{0,t}(1-\hat{p}_{0,t})\sqrt{\frac{\lambda}{2-\lambda}(1-(1-\lambda)^{2t})}$ where $\lambda$ controls how much weight is given to more recent data as opposed to older data, and $\lambda = 0.2$ is recommended by the authors. Also, when the test statistic of the conventional EWMA chart is $Z_{t} > \hat{p}_{0,t} + 0.5L\sigma_{Z_{t}}$, ECDD will report a concept drift warning; when $Z_{t} > \hat{p}_{0,t} + L\sigma_{Z_{t}}$, ECDD will report a concept drift. The control limits $L$ is given by the authors through experimental evaluation.

\begin{figure}[!t]
\centering
\includegraphics[width=0.48\textwidth]{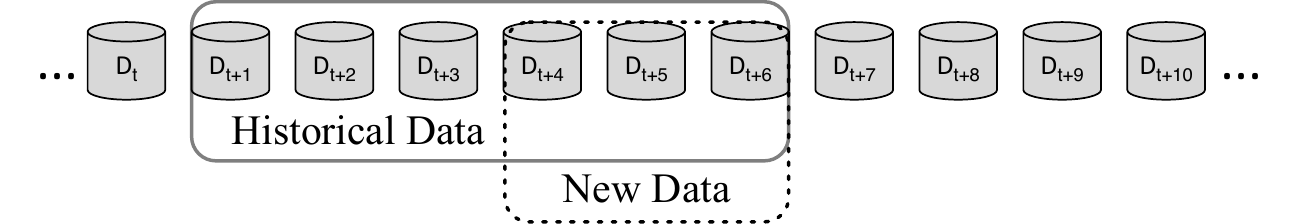}
\caption{Two time windows for concept drift detection. The New Data window has to be defined by the user.}
\label{fig-two-time-window}
\end{figure}

In contrast to DDM and other similar algorithms, Statistical Test of Equal Proportions Detection (STEPD) \cite{N:STEPD} detects error rate change by comparing the most recent time window with the overall time window, and for each timestamp, there are two time windows in the system, as shown in \figurename~\ref{fig-two-time-window}. The size of the new window must be defined by the user. According to \cite{N:STEPD}, the test statistic $\theta_{\text{STEPD}}$ conforms to standard normal distribution, denoted as $\theta _{\text{STEPD}} \sim \mathcal{N}(0,1)$. The significance level of the warning level and the drift level were suggested as $\alpha_{w} = 0.05$ and $\alpha_{d} = 0.003$ respectively. As a result, the warning threshold and drift threshold can be easily calculated.

Another popular two-time window-based drift detection algorithm is \textbf{AD}aptive \textbf{WIN}dowing (ADWIN) \cite{Bifet:ADWIN}. Unlike STEPD, ADWIN does not require users to define the size of the compared windows in advance; it only needs to specify the total size $n$ of a ``sufficiently large'' window $W$. It then examines all possible cuts of $W$ and computes optimal sub-window sizes $n_\text{hist}$ and $n_\text{new}$ according to the rate of change between the two sub-windows $w_\text{hist}$ and $w_\text{new}$. The test statistic is the difference of the two sample means $\theta_{\text{ADWIN}}=|\hat{\mu}_\text{hist}-\hat{\mu}_\text{new}|$. An optimal cut is found when the difference exceeds a threshold with a predefined confidence interval $\delta$. The author proved that both the false positive rate and false negative rate are bounded by $\delta$. It is worth noting that many concept drift adaptation methods/algorithms in the literature are derived from or combined with ADWIN, such as \cite{Bifet:ADWIN_HWT, Bifet:ADWIN_Bag, Bifet:ADWIN_ASHT, Bifet:ADWIN_ARF}. Since their drift detection methods are implemented with almost the same strategy, we will not discuss them in detail.

\subsubsection{Data Distribution-based Drift Detection}
\label{sss-distribution}

\begin{figure}[!t]
\centering
\includegraphics[width=0.48\textwidth]{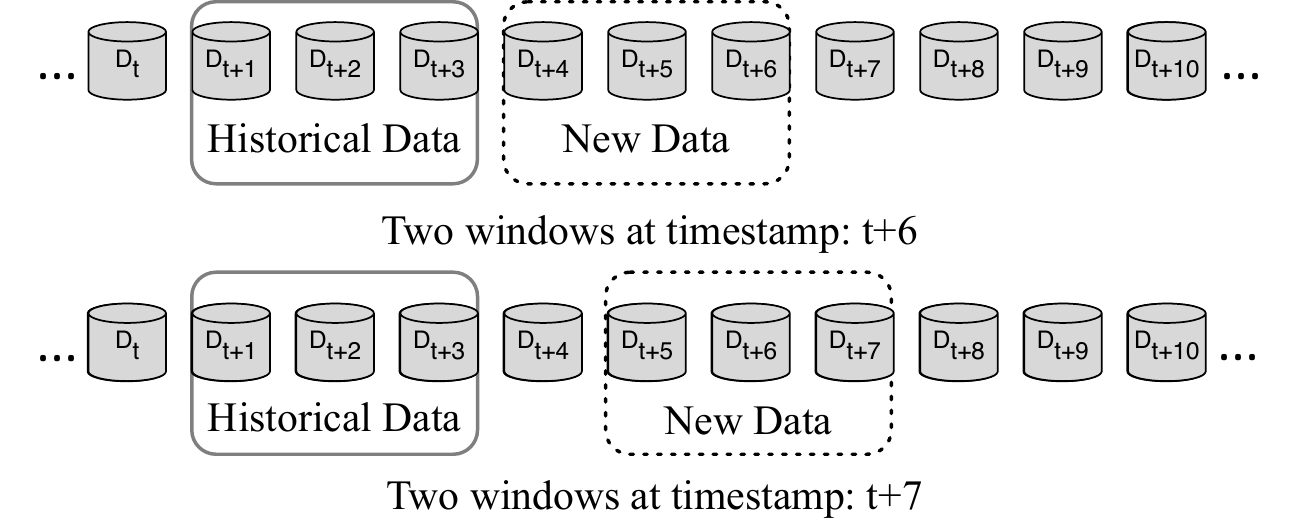}
\caption{Two sliding time windows, of fixed size. The Historical Data window will be fixed while the New Data window will keep moving.}
\label{fig-two-windows}
\end{figure}

The second largest category of drift detection algorithms is data distribution-based drift detection. Algorithms of this category use a distance function/metric to quantify the dissimilarity between the distribution of historical data and the new data. If the dissimilarity is proven to be statistically significantly different, the system will trigger a learning model upgradation process. These algorithms address concept drift from the root sources, which is the distribution drift. Not only can they accurately identify the time of drift, they can also provide location information about the drift. However, these algorithms are usually reported as incurring higher computational cost than the algorithms mentioned in Section~\ref{sss-error-rate} \cite{L:AI2}. In addition, these algorithms usually require users to predefine the historical time window and new data window. The commonly used strategy is two sliding windows with the historical time window fixed while sliding the new data window \cite{D:kdqTree, L:competence, J:PCAdriftDetection}, as shown in \figurename~\ref{fig-two-windows}.

According to the literature, the first formal treatment of change detection in data streams was proposed by \cite{Kifer:RD}. In their study, the authors point out that the most natural notion of distance between distributions is total variation, as defined by: $TV(P_{1},P_{2}) = 2sup_{E\in \varepsilon}|P_{1}(E)-P_{2}(E)|$ or equivalently, when the distribution has the density functions $f_{1}$ and $f_{2}$, $dist_{L^{1}}=\int |f_{1}(x)-f_{2}(x)|\mathrm{d}x$. This provides practical guidance on the design of a distance function for distribution discrepancy analysis. Accordingly, \cite{Kifer:RD} proposed a family of distances, called Relativized Discrepancy (RD). The authors also present the significance level of the distance according to the number of data instances. The bounds on the probabilities of missed detections and false alarms are theoretically proven, using Chernoff bounds and the Vapnik-Chervonenkis dimension. The authors of \cite{Kifer:RD} do not propose novel high-dimensional friendly data models for Stage 2 (data modeling); instead, they stress that a suitable model choice is an open question.

Another typical density-based drift detection algorithm is the Information-Theoretic Approach (ITA) \cite{D:kdqTree}. The intuitive idea underlying this algorithm is to use kdqTree to partition the historical and new data (multi-dimensional) into a set of bins, denoted as $\mathcal{A}$,and then use Kullback-Leibler divergence to quantify the difference of the density $\theta_{\text{ITA}}$ in each bin. The hypothesis test applied by ITA is bootstrapping by merging $W_\text{hist}$, $W_\text{new}$ as $W_\text{all}$ and resampling as ${W'}_\text{hist}$, ${W'}_{new}$ to recompute the $\theta_{\text{ITA}}^{*}$. Once the estimated probability $P(\theta _{\text{ITA}}^{*}\geq \theta _{\text{ITA}}) < 1-\alpha$, concept drift is confirmed, where $\alpha$ is the significant level controlling the sensitivity of drift detection.

Similar distribution-based drift detection methods/algorithms are: Statistical Change Detection for multi-dimensional data (SCD) \cite{song2007statistical}, Competence Model-based drift detection (CM) \cite{L:AI2}, a prototype-based classification model for evolving data streams called SyncStream \cite{J:PCAdriftDetection}, \hly{PCA-based change detection framework (PCA-CD)} \cite{Abdulhakim:PCA}, Equal Density Estimation (EDE) \cite{Feng:EDE}, Least Squares Density Difference-based Change Detection Test (LSDD-CDT) \cite{Alippi:LSDD-CDT}, Incremental version of LSDD-CDT (LSDD-INC) \cite{Alippi:LSDD-INC} and Local Drift Degree-based Density Synchronized Drift Adaptation (LDD-DSDA) \cite{Liu:IJCAI2017}.

\subsubsection{Multiple Hypothesis Test Drift Detection}
\label{sss-multi-hypothesis}

Multiple hypothesis test drift detection algorithms apply similar techniques to those mentioned in the previous two categories. The novelty of these algorithms is that they use multiple hypothesis tests to detect concept drift in different ways. These algorithms can be divided into two groups: 1) parallel multiple hypothesis tests; and 2) hierarchical multiple hypothesis tests.

\begin{figure}[!t]
\centering
\includegraphics[width=0.48\textwidth]{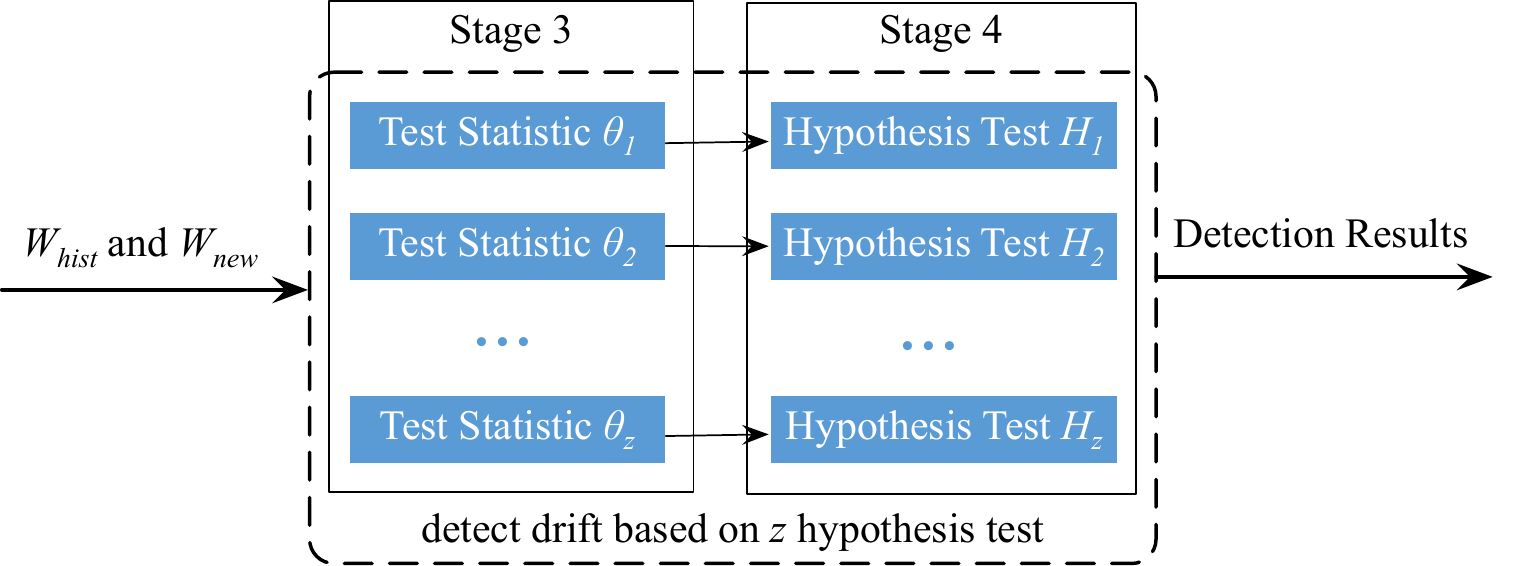}
\caption{Parallel multiple hypothesis test drift detection.}
\label{fig-multi-target-drift-detection}
\end{figure}

The idea of parallel multiple hypothesis drift detection algorithm is demonstrated in \figurename~\ref{fig-multi-target-drift-detection}. According to the literature, Just-In-Time adaptive classifiers (JIT) \cite{Alippi:JITpart1} is the first algorithm that set multiple drift detection hypothesis in this way. The core idea of JIT is to extend the CUSUM chart, known as the Computational Intelligence-based \textbf{CUSUM} test (CI-CUSUM), to detect the mean change in the features interested by learning systems. The authors of \cite{Alippi:JITpart1}, gave the following four configurations for the drift detection target. Config1: the features extracted by Principal Component Analysis (PCA), which removes eigenvalues whose sum is below a threshold, e.g. 0.001. Config2: PCA extracted features plus one generic component of the original features $x_{i}$; Config3: detects the drift in each $x_{i}$ individually. Config4: detects drift in all possible combinations of the feature space $x_{i}$. The authors stated that Config2 is the preferred setting for most situations, according to their experimentation, and also mentioned that Config1 may have a high missing rate, Config3 suffers from a high false alarm rate, and Config4 has exponential computational complexity. The same drift detection strategy has also been applied in \cite{Alippi:JITpart2, Alippi:JIT_NN, Alippi:JIT_ensemble, Alippi:JustInTimeRecurrentDrift} for concept drift adaptation.

Similar implementations have been applied in Linear Four Rate drift detection (LFR) \cite{Heng:LFR}, which maintains and tracks the changes in True Positive rate (TP), True Negative rate (TN), False Positive rate (FP) and False Negative rate (FN) in an online manner. The drift detection process also includes warning and drift levels.

Another parallel multiple hypothesis drift detection algorithm is three-layer drift detection, based on Information Value and Jaccard similarity (IV-Jac) \cite{PeiPei:ThreeLayer}. IV-Jac aims to individually address the label drift $\Prob_{\T}(\y)$ Layer I, feature space drift $\Prob_{\T}(\X)$ Layer II, and decision boundary drift $\Prob_{\T}(\y|\X)$ Layer III. It extracts the Weight of Evidence (WoE) and Information Value (IV) from the available data and then detects whether a significant change exists between the WoE and IV extracted from $W_\text{hist}$ and $W_\text{new}$ by measuring the contribution to the label for a feature value. The hypothesis test thresholds are predefined parameters $\theta_{\Prob_{\T}(\y)} = \theta_{\Prob_{\T}(\X)} = \theta_{\Prob_{\T}(\X|\y)} = 0.5$ by default, which are chosen empirically.

\hly{Ensemble of Detectors (e-Detector) \cite{du2014selective} proposed to detect concept drift via ensemble of heterogeneous drift detector. The authors consider two drift detectors are homogeneous as if they are equivalent in finding concept drifts, otherwise they are heterogeneous. e-Detector groups homogeneous drift detectors via a diversity measurement, named diversity vector. For each group, it select the one with the smallest coefficient of failure as the base detector to form the ensemble. e-Detector reports concept drift following the early-find-early-report rule, which means no matter which base detector detect a drift, the e-Detector reports a drift.
Similar strategy has been applied in drift detection ensemble (DDE) \cite{maciel2015lightweight}.}

\begin{figure}[!t]
\centering
\includegraphics[width=0.48\textwidth]{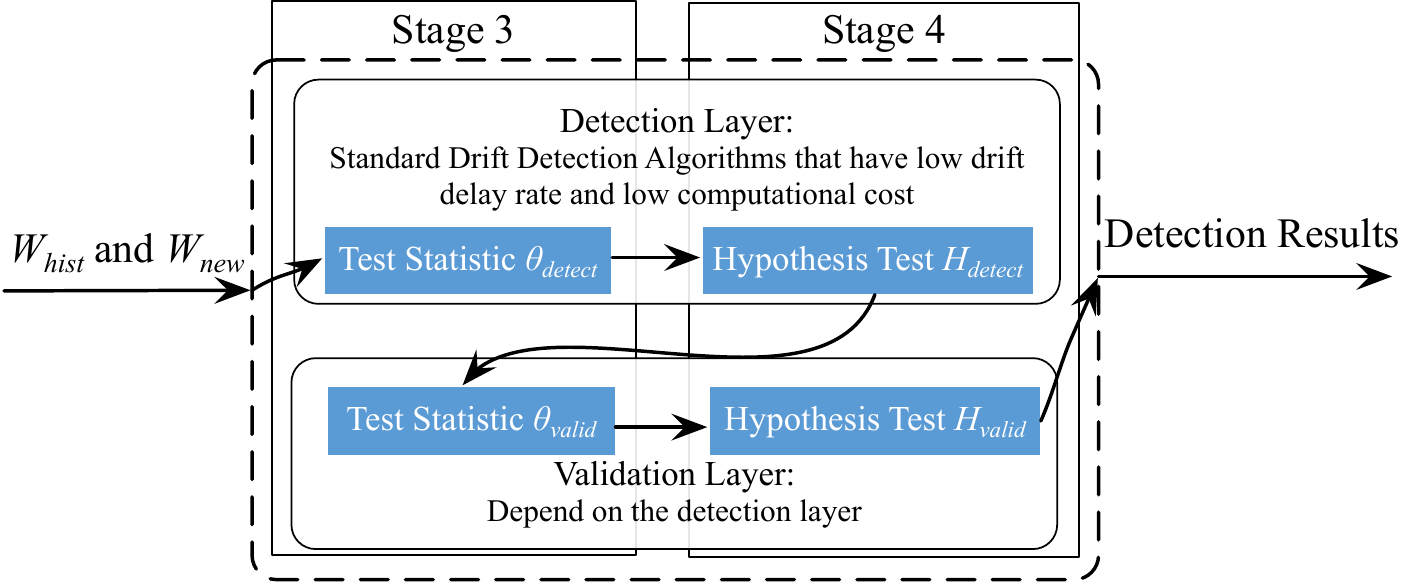}
\caption{Hierarchical multiple hypothesis test drift detection.}
\label{fig-hierarchical-drift-detection}
\end{figure}

Hierarchical drift detection is an emerging drift detection category that has a multiple verification schema. The algorithms in this category usually detect drift using an existing method, called the detection layer, and then apply an extra hypothesis test, called the validation layer, to obtain a second validation of the detected drift in a hierarchical way. The overall workflow is shown in \figurename~\ref{fig-hierarchical-drift-detection}.

According to the claim made by \cite{Alippi:HCDTs}, Hierarchical Change-Detection Tests (HCDTs) is the first attempt to address concept drift using a hierarchical architecture. The detection layer can be any existing drift detection method that has a low drift delay rate and low computational burden. The validation layer will be activated and deactivated based on the results returned by the detection layer. The authors recommend two strategies for designing the validation layer: 1) estimating the distribution of the test statistics by maximizing the likelihood; 2) adapting an existing hypothesis test, such as the Kolmogorov-Smirnov test or the Cramer-Von Mises test.

Hierarchical Linear Four Rate (HLFR) \cite{Yu:HLFR} is another recently proposed hierarchical drift detection algorithm. It applies the drift detection algorithm LFR as the detection layer. Once a drift is confirmed by the detection layer, the validation layer will be triggered. The validation layer of HLFR is simply a zero-one loss, denoted as $E$, over the ordered train-test split. If the estimated zero-one loss exceeds a predefined threshold, $\eta =0.01$, the validation layer will confirm the drift and report to the learning system to trigger a model upgradation process.

Two-Stage Multivariate Shift-Detection based on \textbf{EWMA} (TSMSD-EWMA) \cite{Haider:PR_EWMA} has a very similar implementation, however, the authors do not claim that their method is a hierarchy-based algorithm.

\hly{Hierarchical Hypothesis Testing with Classification Uncertainty (HHT-CU) and Hierarchical Hypothesis Testing with Attribute-wise "Goodness-of-fit" (HHT-AG) are two drift detection algorithms based on request and reverify strategy \cite{yu2018request}. For HHT-CU, the detection layer is a hypotheses test based on Heoffding's inequality that monitoring the change of the classification uncertainty measurement. The validation layer is a permutation test that evaluates the change of the zero-one loss of the learner. For HHT-AG, the detection layer is conducted based on Kolmogorov-Smirnov (KS) test for each feature distribution. Then HHT-AG validate the potential drift points by requiring true labels of data that come from $w_{new}$, and performing $d$ independent two-dimensional (2D) KS test with each feature-label bivariate distribution.
Compare to other drift detection algorithms, HHT-AG can handle concept drift with less true labels, which makes it more powerful when dealing with high verification latency.}

\subsection{Summary of concept drift detection methods/algorithms}
\label{sub-dd-summary}

\tablename~\ref{tab-drift-detection} lists the most popular concept drift detection methods/algorithms against the general framework summarized in Section~\ref{sub-dd-framework} (\figurename~\ref{fig-framework-drift-detection}). \hly{A comparative study on eight popular drift detection methods can be found in \cite{gonccalves2014comparative}.}


\begin{table*}[!t]
\renewcommand{\arraystretch}{1.1}
\caption{Summary of drift detection algorithms}
\label{tab-drift-detection}
\centering
\scriptsize
\begin{tabular}{|p{1.3cm}|p{2.3cm}|p{3.5cm}|p{2.5cm}|p{3cm}|p{3cm}|}
\hline
Category & Algorithms & Stage 1 & Stage 2 & Stage 3 & Stage 4 \\
\hline
\multirow{1}{1.3cm}{Error rate-based} & DDM\cite{Gama:DDM} & Landmark & Learner & Online error rate & Distribution estimation \\
\cline{2-6}                           & EDDM\cite{B:EDDM} & Landmark  & Learner & Online error rate & Distribution estimation \\
\cline{2-6}                           & FW-DDM\cite{Liu:FuzzyWindow} & Landmark & Learner & Online error rate & Distribution estimation \\
\cline{2-6}                           & DEML\cite{Xu:DELM} & Landmark & Learner & Online error rate & Distribution estimation \\
\cline{2-6}                           & STEPD\cite{N:STEPD} & Predefined $w_\text{hist}$, $w_\text{new}$ & Learner & Error rate difference & Distribution estimation \\
\cline{2-6}                           & ADWIN\cite{Bifet:ADWIN} & Auto cut $w_\text{hist}$, $w_\text{new}$ & Learner & Error rate difference & Hoeffding's Bound \\
\cline{2-6}                           & ECDD\cite{G:ECDD} & Landmark & Learner & Online error rate & EWMA Chart \\
\cline{2-6}                           & HDDM\cite{I:HDDM} & Landmark & Learner & Online error rate & Hoeffding's Bound \\
\cline{2-6}                           & LLDD \cite{Gama:local} & Landmark, or sliding $w_\text{hist}$, $w_\text{new}$ & Decision trees & Tree node error rate & Hoeffding's Bound \\
\hline
\multirow{1}{1.3cm}{Data distribution-based} & kdqTree\cite{D:kdqTree} & Fixed $w_\text{hist}$, Sliding $w_\text{new}$ & kdqTree & KL divergence & Bootstrapping \\
\cline{2-6}                                  & CM\cite{L:AI2, L:competence} & Fixed $w_\text{hist}$, Sliding $w_\text{new}$ & Competence model & Competence distance & Permutation test \\
\cline{2-6}                                  & RD \cite{Kifer:RD} & Fixed $w_\text{hist}$, Sliding $w_\text{new}$ & KS structure & Relativized Discrepancy & VC-Dimension \\
\cline{2-6}                                  & SCD \cite{song2007statistical} & Fixed $w_\text{hist}$, Sliding $w_\text{new}$ & kernel density estimator & log-likelihood & Distribution estimation \\
\cline{2-6}                                  & EDE\cite{Feng:EDE} & Fixed $w_\text{hist}$, Sliding $w_\text{new}$ & Nearest neighbor & Density scale & Permutation test \\
\cline{2-6}                                  & SyncStream \cite{J:PCAdriftDetection} & Fixed $w_\text{hist}$, Sliding $w_\text{new}$ & PCA & P-Tree & Wilcoxon test \\
\cline{2-6}                                  & \hly{PCA-CD \cite{Abdulhakim:PCA}} & \hly{Fixed $w_\text{hist}$, Sliding $w_\text{new}$} & \hly{PCA} & \hly{Change-Score} & \hly{Page-Hinkley test} \\
\cline{2-6}                                  & LSDD-CDT\cite{Alippi:LSDD-CDT} & Fixed $w_\text{hist}$, Sliding $w_\text{new}$ & Learner & Relative difference & Distribution estimation \\
\cline{2-6}                                  & LSDD-INC\cite{Alippi:LSDD-INC} & Fixed $w_\text{hist}$, Sliding $w_\text{new}$ & Learner & Relative difference & Distribution estimation \\
\cline{2-6}                                  & LDD-DSDA\cite{Liu:IJCAI2017} & Fixed $w_\text{hist}$, Sliding $w_\text{new}$ & k-nearest neighbor & Local drift degree & Distribution estimation \\
\hline
\multirow{1}{1.3cm}{Multiple Hypothesis Tests}  & JIT\cite{Alippi:JITpart1} & Landmark & Selected features & 4 configurations & Distribution estimation \\
\cline{2-6}                                     & LFR\cite{Heng:LFR} & Landmark & Learner & TP, TN, FP, FN & Distribution estimation \\
\cline{2-6}                                     & Three-layer\cite{PeiPei:ThreeLayer} & Sliding both $w_\text{hist}$, $w_\text{new}$ & Learner & $\Prob(\y)$, $\Prob(\X)$, $\Prob(\X|\y)$ & Distribution estimation \\
\cline{2-6}                                     & \hly{e-Detector\cite{du2014selective}} & \hly{depends on base detector} & \hly{depends} & \hly{depends} & \hly{depends} \\
\cline{2-6}                                     & \hly{DDE\cite{maciel2015lightweight}} & \hly{depends on base detector} & \hly{depends} & \hly{depends} & \hly{depends} \\
\cline{2-6}                                     & TSMSD-EWMA\cite{Haider:PR_EWMA} & Landmark & Learner & Online error rate & EWMA Chart \\
\cline{2-6}                                     & HCDTs\cite{Alippi:HCDTs} & Landmark & Depending on layers & Depending on layers & Depending on layer \\
\cline{2-6}                                     & HLFR\cite{Yu:HLFR} & Landmark & Learner & TP, TN, FP, FN & Distribution estimation \\
\cline{2-6}                                     & \hly{HHT-CU\cite{yu2018request}} & \hly{Landmark} & \hly{Learner} & \hly{Classification uncertainty} & \hly{Layer-I Hoeffding's Bound, Layer-II Permutation Test} \\
\cline{2-6}                                     & \hly{HHT-AG\cite{yu2018request}} & \hly{Fixed $w_\text{hist}$, Sliding $w_\text{new}$} & \hly{N/A} & \hly{KS statistic on each attribute} & \hly{Layer-I KS test, Layer -II 2D KS test} \\
\hline
\end{tabular}

\end{table*}


\section{Concept Drift understanding}
\label{sec-du}

Drift understanding refers to retrieving concept drift information about ``When'' (the time at which the concept drift occurs and how long the drift lasts), ``How'' (the severity /degree of concept drift), and ``Where'' (the drift regions of concept drift). This status information is the output of the drift detection algorithms, and is used as input for drift adaptation.

\subsection{The time of concept drift occurs (When)}
\label{sub-du-when}

The most basic function of drift detection is to identify the timestamp when a drift occurs. Recalling the definition of concept drift $\exists \T\colon \Prob_{\T}(\X,\y)\neq \Prob_{\T+1}(\X,\y)$, the variable $\T$ represents the time at which a concept drift occurs. In drift detection methods/algorithms, an alarm signal is used to indicate whether the concept drift has or has not occurred or not at the current timestamp. It is also a signal for a learning system to adapt to a new concept. Accurately identifying the time a drift occurs is critical to the adaptation process of a learning system; a delay or a false alarm will lead to failure of the learning system to track new concepts.

A drift alarm usually has a statistical guarantee with a predefined false alarm rate. Error rate-based drift detection algorithms monitor the performance of the learning system, based on statistical process control. For example, DDM \cite{Gama:DDM} sends a drift signal when the learning accuracy of the learner drops below a predefined threshold, which is chosen by the three-sigma rule \cite{pukelsheim1994three}. ECCD \cite{G:ECDD} reports a drift when the online error rate exceeds the control limit of EWMA. Most data distribution-based drift detection algorithms report a drift alarm when two data samples have a statistically significant difference. PCA-based drift detection \cite{J:PCAdriftDetection} outputs a drift signal when the $p$-value of the generalized Wilcoxon test statistic $W_{BF}^{l}$ is significantly large. The method in \cite{L:competence} confirms that a drift has occurred by verifying whether the empirical competence-based distance is significantly large through permuataion test.

\begin{figure}[!t]
\centering
\includegraphics[width=0.48\textwidth]{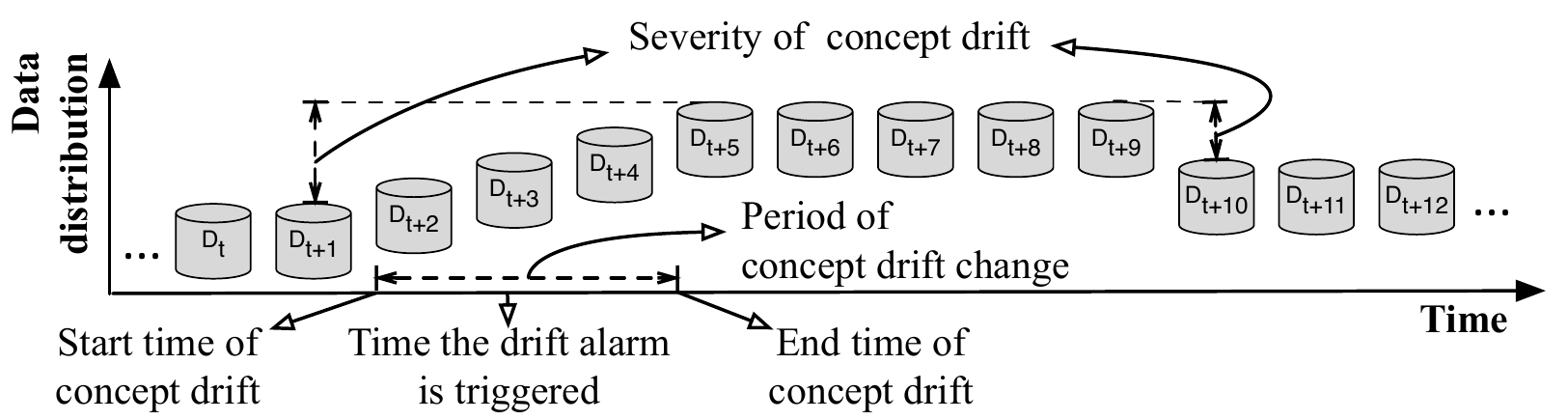}
\caption{An example of the occurrence time and the severity of concept drift. One incremental drift starts to change at $\T+1$ and ends at $\T+5$. The other sudden drift occurs between $\T+9$ and $\T+10$. The severity of these two concept drifts ($D_{\T+1}$-$D_{\T+5}$ and $D_{\T+9}$-$D_{\T+10}$) is marked with brackets.}
\label{fig-drift-understanding}
\end{figure}

Taking into account the various drift types, concept drift understanding needs to explore the start time point, the change period, and the end time point of concept drift. And these time information could be useful input for the adaptation process of the learning system. However the drift timestamp alert in existing drift detection algorithms is delayed compared to the actual drifting timestamp, since most drift detectors require a minimum number of new data to evaluate the status of the drift, as shown in \figurename~\ref{fig-drift-understanding}. The emergence time of the new concept is therefore still vague. Some concept drift detection algorithms such as DDM \cite{Gama:DDM}, EDDM \cite{B:EDDM}, STEPD\cite{N:STEPD}, and HDDM \cite{I:HDDM}, trigger a warning level to indicate a drift may have occurred. The threshold used to trigger warning level is a relaxed condition of the threshold used for the drift level; for example, the warning level is set $p$-value to 95\% or $2\sigma$, and the drift level is set $p$-value to 99\% or $3\sigma$. The data accumulated between the warning level and the drift level are used as the training set for updating a learning model.

\subsection{The severity of concept drift (How)}
\label{sub-du-how}

The severity of concept drift refers to using a quantified value to measure the similarity between the new concept and the previous concept, as shown in \figurename~\ref{fig-drift-understanding}. Formally, the severity of concept drift can be represented as $\Delta = \delta (\Prob_{\T}(\X,\y), \Prob_{\T+1}(\X,\y))$, where $\delta$ is a function to measure the discrepancy of two data distributions, and $\T$ is the timestamp when the concept drift occurred.
\hly{$\Delta$ usually is a non-negative value indicating the severity of concept drift. The greater the value of $\Delta$, the larger the severity of the concept drift is.}


In general, error rate-based drift detection cannot directly measure the severity of concept drift, because it mainly focuses on monitoring the performance of the learning system, not the changes in the concept itself. However, the degree of decrease in learning accuracy can be used as an indirect measurement to indicate the severity of concept drift. If learning accuracy has dropped significantly when drift is observed, this indicates that the new concept is different from the previous one. For example, the severity of concept drift could be reflected by the difference between $p_{i}$ and $p_\text{min}$ in \cite{Gama:DDM, Xu:DELM}\hly{, denoted as $\Delta \sim p_{i} - p_\text{min}$}; the difference between overall accuracy $\hat{p}_{\text{hist}}$ and recent accuracy $\hat{p}_{\text{new}}$ in \cite{N:STEPD}\hly{, expressed as $\Delta \sim \hat{p}_{\text{new}} - \hat{p}_{\text{hist}}$}; and the difference between test statistics in the former window $E[\hat{X}_\text{cut}]$ and test statistics in the later window $E[\hat{Y}_{i-\text{cut}}]$ \cite{I:HDDM}, \hly{denoted as $\Delta \sim E[\hat{Y}_{i-\text{cut}}] - E[\hat{X}_\text{cut}]$}. However, the meaning of these differences is not discussed in existing publications. The ability of error rate-based drift detection to output the severity of concept drift is still vague.

Data distribution-based drift detection methods can directly quantify the severity of concept drift since the measurement used to compare two data samples already reflects the difference. For example, \cite{Kifer:RD} employed \hly{\emph{a relaxation of the total variation distance $d_\mathcal{A}(S_1, S_2)$}} to measure the difference between two data distributions. \cite{L:competence} proposed \hly{\emph{a competence-based empirical distance $d^{CB}(S_1, S2)$}} to show the difference between two data samples. Other drift detection methods have used information-theoretic distance; for example, \hly{\emph{Kullback-Leibler divergence $D(W_1 || W_2)$}}, also called relative entropy, was used in the study reported in \cite{D:kdqTree}. The range of these distances is $[0,1]$. The greater the distance, the larger the severity of the concept drift is. The distance ``1'' means that a new concept is different from the pervious one, while the distance ``0'' means that two data concepts are identical. The \hly{\emph{test statistic $\delta$}} used in \cite{song2007statistical} gives an extremely small negative value if the new concept is quite different from the previous concept. The degree of concept drift in \cite{J:PCAdriftDetection} is measured by the resulting $p$-value of \hly{the test statistic $W^l_{BF}$ and the defined $Angle(D_t, D_{t+1})$ of two datasets $D_t$ and $D_{t+1}$}.

The severity of concept drift can be used as a guideline for choosing drift adaptation strategies. For example, if the severity of drift in a classification task is low, the decision boundary may not move much in the new concept. Thus, adjusting the current learner by incremental learning will be adequate.  In contrast, if the severity of the concept drift is high, the decision boundary could change significantly, therefore discarding the old learner and retraining a new one could be better than incrementally updating the old learner. We would like to mention that, even though some researches have promoted the ability to describe and quantify the severity of the detected drift, this information is not yet widely utilized in drift adaptation.

\subsection{The drift regions of concept drift (Where)}
\label{sub-du-where}

\begin{figure}[!t]
\centering
\subfloat[Data at time $\T$]{\includegraphics[width=0.16\textwidth]{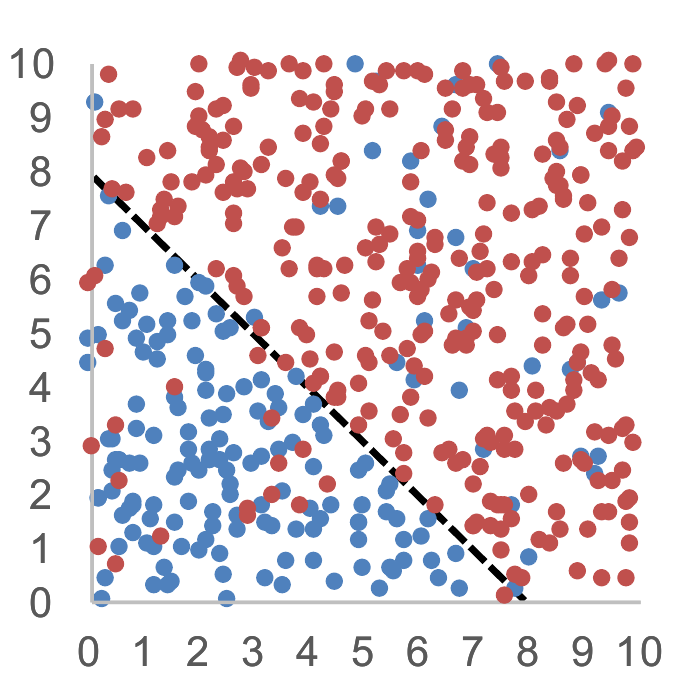}%
\label{fig-drift-regions-left}}
\subfloat[Data at time $\T+1$]{\includegraphics[width=0.16\textwidth]{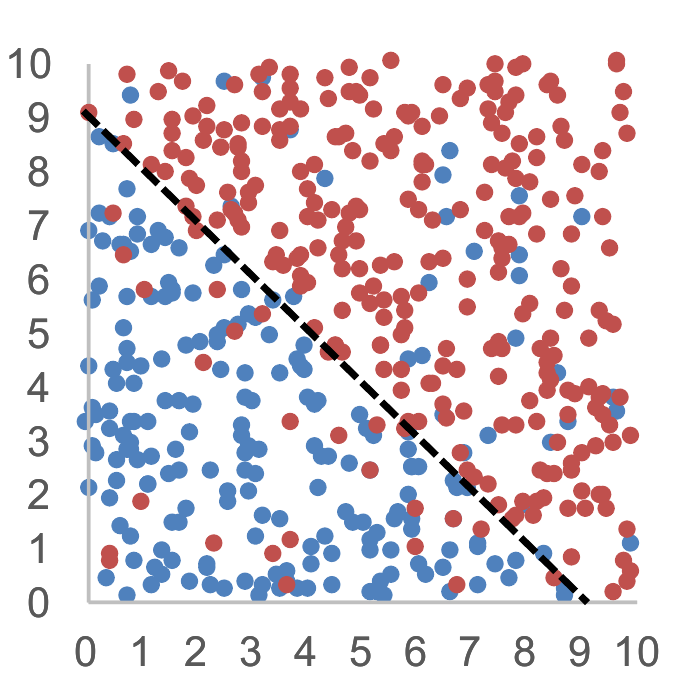}%
\label{fig-drift-regions-middle}}
\subfloat[Data in drift regions]{\includegraphics[width=0.16\textwidth]{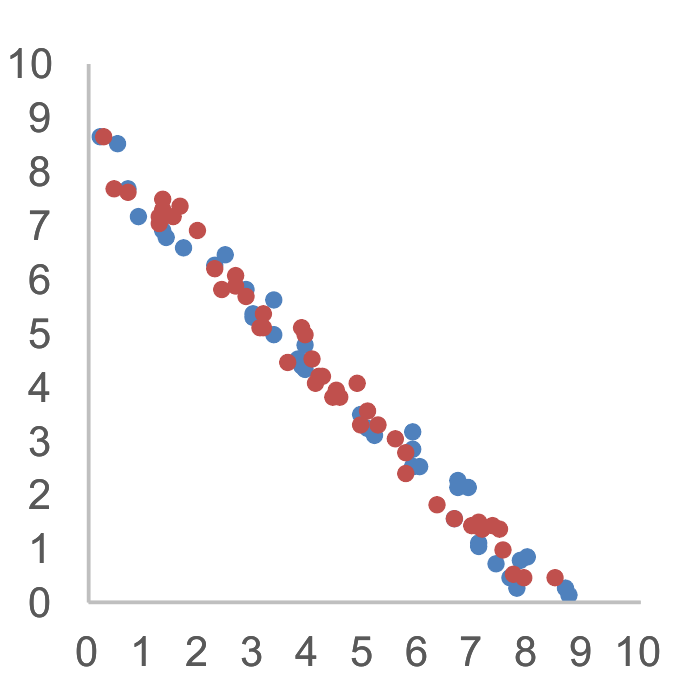}%
\label{fig-drift-regions-right}}
\caption{An example of the drift regions of concept drift.}
\label{fig-drift-regions}
\end{figure}

The drift regions of concept drift are the regions of conflict between a new concept and the previous concept. Drift regions are located by finding regions in \hly{data feature space $\X$} where $\Prob_{\T}(\X,\y)$ and $\Prob_{\T+1}(\X,\y)$ have significant difference. To illustrate this, we give an example of a classification task in \figurename~\ref{fig-drift-regions}. The data used in this task are uniformly sampled in the range of $[0,10]^{2}$. The left sub-figure is the data sample accumulated at time $\T$, where the decision boundary is $x_{1}+x_{2}=8$. The middle sub-figure is the data accumulated at time $\T+1$, where the decision boundary is $x_{1}+x_{2}=9$. Intuitively, data located in regions $8\leq x_{1}+x_{2}<9$ have different classes in time $\T$ and time $\T+1$, since the decision boundary has changed. The right sub-figure shows the data located in the drift regions.

The techniques to identify drift regions are highly dependent on the data model used in the drift detection methods/algorithms. Paper \cite{Gama:local} detected drift data in local regions of the instance space by using online error-rate inside the inner-nodes of a decision tree. The whole data feature space is partitioned by decision tree. Each leaf of this decision tree corresponds to a hyper-rectangle in the data feature space. All leaf nodes contain a drift detector. When the leaf nodes are alerted to a drift, the corresponding hyper-rectangles indicate the regions of drift in the data feature space. Similar to \cite{Gama:local}, \cite{D:kdqTree} utilized the nodes of the \textit{kdq-tree} with Kulldorff's spatial scan statistic to identify the regions in which data had changed the most. Once a drift has been reported, Kulldorff's statistic measures how the two datasets differ only with respect to the region associated with the leaf node of the \textit{kdq-tree}. The leaf nodes with the greater statistical value of show the drift regions. \cite{L:AI2} highlighted the most severe regions of concept drift through top-$p$-competence areas. Utilizing the \textit{RelatedSets} of the competence model, the data feature space is partitioned by a set of overlapping hyperspheres. The \textit{RelatedSet}-based empirical distance defines the distance between two datasets on a particular hypersphere. The drift regions are identified by the corresponding hyperspheres with large empirical distance at top $p$\% level. \cite{Liu:IJCAI2017} identified drift regions by monitoring the discrepancy in the regional density of data, which is called the local drift degree. A local region with a corresponding increase or decrease in density will be highlighted as a critical drift region.

Locating concept drift regions benefits drift adaptation. Paper \cite{tsymbal2008dynamic} pointed out that even if the concept of the entire dataset drifts, there are regions of the feature space that will remain stable significantly longer than other regions. In an ensemble scenario, the old learning models of stable regions could still be useful for predicting data instances located within stable regions, or data instances located in drift regions could be used to build a more updated current model. The authors of \cite{L:competence} also pointed out that identifying drift regions can help in recognizing obsolete data that conflict with current concepts and distinguish noise data from novel data. In their later research \cite{L:AI2}, they utilized top-$p$-competence areas to edit cases in a case-based reasoning system for fast new concept switching. One step in their drift adaptation is to remove conflict instances. To preserve as many instances of a new concept as possible, they only remove obsolete conflict instances which are outside the drift regions.  LDD-DSDA \cite{Liu:IJCAI2017} utilized drift regions as an instance selection strategy to construct a training set that continually tracked a new concept.

\subsection{Summary of drift understanding}

We summarize concept drift detection algorithms according to their ability to identify when, how, and where concept drift occurs, as shown in \tablename~\ref{tab-drift-understanding}. All drift detection algorithms can identify the occurrence time of concept drift (when), and most data distribution-based drift detection algorithms can also measure the severity of concept drift (how) through the test statistics, but only a few algorithms have the ability to locate drift regions (where).


\begin{table}[!t]
\renewcommand{\arraystretch}{1.1}
\caption{Summary of drift understanding for drift detection algorithms}
\label{tab-drift-understanding}
\centering
\scriptsize
\begin{tabular}{|p{1.8cm}|l|p{0.5cm}|p{0.45cm}|p{0.55cm}|}
\hline
Category & Algorithms & When & How & Where \\
\hline
\multirow{1}{2cm}{Error rate-based}         & DDM \cite{Gama:DDM}                   & $\surd$ &         &           \\
\cline{2-5}                                 & EDDM \cite{B:EDDM}                    & $\surd$ &         &           \\
\cline{2-5}                                 & FW-DDM \cite{Liu:FuzzyWindow}         & $\surd$ &         &           \\
\cline{2-5}                                 & DEML \cite{Xu:DELM}                   & $\surd$ &         &           \\
\cline{2-5}                                 & STEPD \cite{N:STEPD}                  & $\surd$ &         &           \\
\cline{2-5}                                 & ADWIN \cite{Bifet:ADWIN}              & $\surd$ &         &           \\
\cline{2-5}                                 & ECDD \cite{G:ECDD}                    & $\surd$ &         &           \\
\cline{2-5}                                 & HDDM \cite{I:HDDM}                    & $\surd$ &         &           \\
\cline{2-5}                                 & LLDD \cite{Gama:local}                & $\surd$ &         & $\surd$   \\
\hline
\multirow{1}{2cm}{Data distribution-based}  & kdqTree \cite{D:kdqTree}              & $\surd$ & $\surd$ & $\surd$   \\
\cline{2-5}                                 & CM  \cite{L:AI2, L:competence}        & $\surd$ & $\surd$ & $\surd$   \\
\cline{2-5}                                 & RD  \cite{Kifer:RD}                   & $\surd$ & $\surd$ &           \\
\cline{2-5}                                 & SCD \cite{song2007statistical}        & $\surd$ & $\surd$ &           \\
\cline{2-5}                                 & EDE \cite{Feng:EDE}                   & $\surd$ &         &           \\
\cline{2-5}                                 & SyncStream \cite{J:PCAdriftDetection} & $\surd$ & $\surd$ &           \\
\cline{2-5}                                 & \hly{PCA-CD} \cite{Abdulhakim:PCA}    & $\surd$ & $\surd$ &           \\
\cline{2-5}                                 & LSDD-CDT \cite{Alippi:LSDD-CDT}       & $\surd$ &         &           \\
\cline{2-5}                                 & LSDD-INC \cite{Alippi:LSDD-INC}       & $\surd$ &         &           \\
\cline{2-5}                                 & LDD-DSDA \cite{Liu:IJCAI2017}         & $\surd$ & $\surd$ & $\surd$   \\
\hline
\multirow{1}{2cm}{Multiple hypothesis tests}& JIT \cite{Alippi:JITpart1}            & $\surd$ &         &           \\
\cline{2-5}                                 & LFR \cite{Heng:LFR}                   & $\surd$ &         &           \\
\cline{2-5}                                 & Three-layer drift detection \cite{PeiPei:ThreeLayer} & $\surd$ & &    \\
\cline{2-5}                                 & \hly{e-Detector \cite{du2014selective}}  & $\surd$ &      &           \\
\cline{2-5}                                 & \hly{DDE \cite{maciel2015lightweight}}   & $\surd$ &      &           \\
\cline{2-5}                                 & EWMA \cite{Haider:PR_EWMA}            & $\surd$ &         &           \\
\cline{2-5}                                 & HCDTs \cite{Alippi:HCDTs}             & $\surd$ &         &           \\
\cline{2-5}                                 & HLFR \cite{Yu:HLFR}                   & $\surd$ &         &           \\
\cline{2-5}                                 & \hly{HHT-CU\cite{yu2018request}}      & $\surd$ &         &           \\
\cline{2-5}                                 & \hly{HHT-AG\cite{yu2018request}}      & $\surd$ &         &           \\

\hline
\end{tabular}

\end{table}

\section{Drift adaptation}
\label{sec-ka}

This section focuses on strategies for updating existing learning models
according to the drift, which is known as drift adaptation or reaction.
There are three main groups of drift adaptation methods, namely simple
  retraining, ensemble retraining and model adjusting, that aim to handle
  different types of drift.

\subsection{Training new models for global drift}
\label{sub-ka-retraining}

\begin{figure}[!t]
\centering
\includegraphics[width=0.48\textwidth]{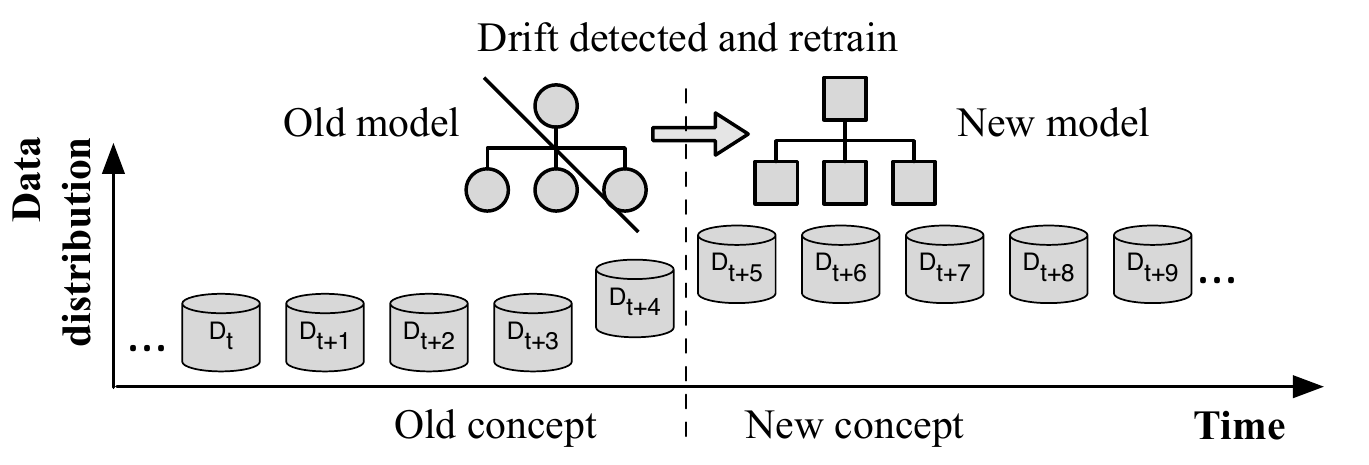}
\caption{A new model is trained with latest data to replace the old model when
  a concept drift is detected.}
\label{fig-drift-adaptation-retrain}
\end{figure}

Perhaps the most straightforward way of reacting to concept drift is to retrain
a new model with the latest data to replace the obsolete model, as shown in
\figurename~\ref{fig-drift-adaptation-retrain}. An explicit concept drift
detector is required to decide when to retrain the model (see Section 3 on drift
detection). A window strategy is often adopted in this method to preserve the
most recent data for retraining and/or old data for distribution change test.
\textit{Paired Learners} \cite{bach2008paired} follows this strategy and uses
two learners: the \textit{stable learner} and the \textit{reactive learner}. If
the stable learner frequently misclassifies instances that the reactive learner
correctly classifies, a new concept is detected and the stable learner will be
replaced with the reactive learner. This method is simple to understand and easy
to implement, and can be applied at any point in the data stream.

When adopting a window-based strategy, a trade-off must be made in order to
decide an appropriate window size. A small window can better reflect the latest
data distribution, but a large window provides more data for training a new
model. A popular window scheme algorithm that aims to mitigate this problem is
ADWIN \cite{Bifet:ADWIN}. Unlike most earlier works, it does not require the
user to guess a fixed size of the windows being compared in advance; instead, it
examines all possible cuts of the window and computes optimal sub-window sizes
according to the rate of change between the two sub-windows. After the optimal
window cut has been found, the window containing old data is dropped and a new
model can be trained with the latest window data.

Instead of directly retraining the model, researchers have also attempted to
integrate the drift detection process with the retraining process for specific
machine learning algorithms. DELM \cite{Xu:DELM} extends the traditional ELM
algorithm with the ability to handle concept drift by adaptively adjusting the
number of hidden layer nodes. When the classification error rate increases ---
which could indicate the emergence of a concept drift --- more nodes are added
to the network layers to improve its approximation capability. Similarly, FP-ELM
\cite{liu2016fpelm} is an ELM-extended method that adapts to drift by
introducing a forgetting parameter to the ELM model. A parallel version of
ELM-based method \cite{Han2015} has also been developed for high-speed
classification tasks under concept drift. OS-ELM \cite{soares2016an} is another
online learning ensemble of repressor models that integrates ELM using an
ordered aggregation (OA) technique, which overcomes the problem of defining the
optimal ensemble size.

Instance-based lazy learners for handling concept drift have also been studied
intensively. The \textit{Just-in-Time} adaptive classifier
\cite{Alippi:JITpart1, Alippi:JITpart2} is such a method which follows the
"detect and update model" strategy. For drift detection, it extends the
traditional CUSUM test \cite{manly2000a} to a pdf-free form. This detection
method is then integrated with a kNN classifier \cite{Alippi:JITpart2}. When a
concept drift is detected, old instances (more than the last $T$ samples) are
removed from the case base. In later work, the authors of
\cite{Alippi:JustInTimeRecurrentDrift, silva2013data} extended this algorithm to
handle recurrent concepts by computing and comparing current concept to
previously stored concepts. NEFCS \cite{L:AI2} is another kNN-based adaptive
model. A competence model-based drift detection algorithm \cite{L:competence}
was used to locate drift instances in the case base and distinguish them from
noise instances and a redundancy removal algorithm, Stepwise Redundancy Removal
(SRR), was developed to remove redundant instances in a uniform way,
guaranteeing that the reduced case base would still preserve enough information
for future drift detection.

\subsection{Model ensemble for recurring drift}
\label{sub-ka-ensemble}

\begin{figure}[!t]
\centering
\includegraphics[width=0.48\textwidth]{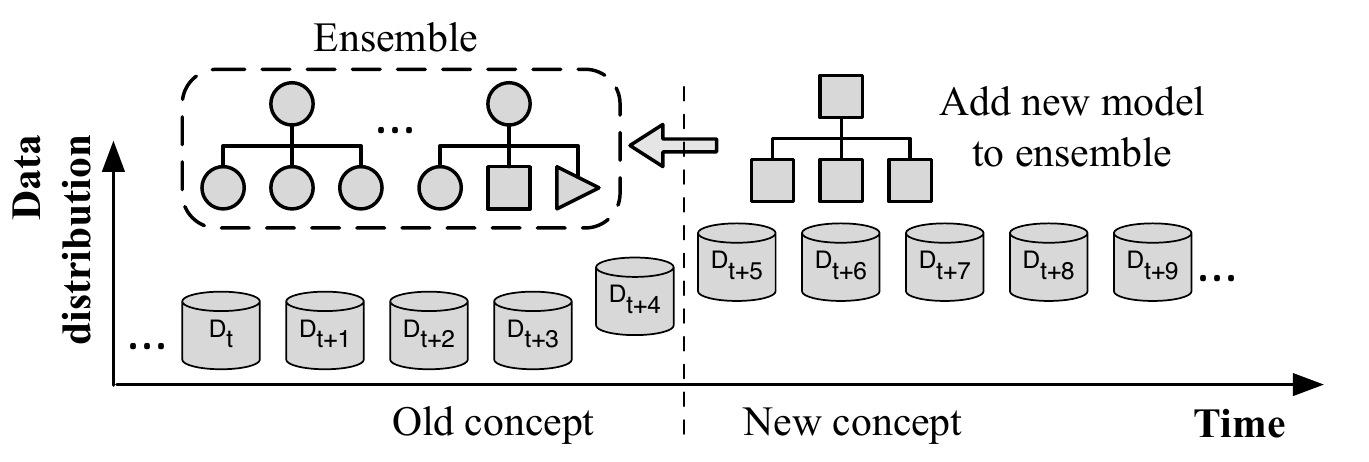}
\caption{A new base classifier is added to the ensemble when a concept drift occurs.}
\label{fig-drift-adaptation-ensemble}
\end{figure}

In the case of recurring concept drift, preserving and reusing old
  models can save significant effort to retrain a new model for recurring
  concepts. This is the core idea of using ensemble methods to handle concept
  drift. Ensemble methods have received much attention in stream data mining
research community in recent years. Ensemble methods comprise a set of base
classifiers that may have different types or different parameters. The output of
each base classifier is combined using certain voting rules to predict the newly
arrived data. Many adaptive ensemble methods have been developed that aim to
handle concept drift by extending classical ensemble methods or by creating
specific adaptive voting rules. An example is shown in
\figurename~\ref{fig-drift-adaptation-ensemble}, where new base classifier is
added to the ensemble when concept drift occurs.

Bagging, Boosting and Random Forests are classical ensemble methods used to
improve the performance of single classifiers. They have all been extended for
handling streaming data with concept drift. An online version of the bagging
method was first proposed in \cite{oza2001experimental} which uses each instance
only once to simulate the batch mode bagging. In a later study
\cite{bifet2010leveragin}, this method was combined with the ADWIN drift
detection algorithm \cite{Bifet:ADWIN} to handle concept drift. When a concept
drift is reported, the newly proposed method, called Leveraging Bagging, trains
a new classifier on the latest data to replace the existing classifier with the
worst performance. Similarly, an adaptive boosting method was developed in
\cite{chu2004fast} which handles concept drift by monitoring prediction accuracy
using a hypothesis test, assuming that classification errors on non-drifting
data should follow Gaussian distribution. In a recent work
\cite{Bifet:ADWIN_ARF}, the Adaptive Random Forest (ARF) algorithm was proposed,
which extends the random forest tree algorithm with a concept drift detection
method, such as ADWIN \cite{Bifet:ADWIN}, to decide when to replace an obsolete
tree with a new one. A similar work can be found in \cite{li2015learning}, which
uses Hoeffding bound to distinguish concept drift from noise within decision
trees.

Besides extending classical methods, many new ensemble methods have been
developed to handle concept drift using novel voting techniques.
\textbf{D}ynamic \textbf{W}eighted \textbf{M}ajority (DWM)
\cite{kolter2007dynamic} is such an ensemble method that is capable of adapting
to drifts with a simple set of weighted voting rules. It manages base
classifiers according to the performance of both the individual classifiers and
the global ensemble. If the ensemble misclassifies an instance, DWM will train a
new base classifier and add it to ensemble. If a base classifier misclassifies
an instance, DWM reduces its weight by a factor. When the weight of a base
classifier drops below a user defined threshold, DWM removes it from the
ensemble. The drawback of this method is that the adding classifier process may
be triggered too frequently, introducing performance issues on some occasions,
such as when gradual drift occurs. A well-known ensemble method, Learn++NSE
\cite{elwell2011incremental}, mitigates this issue by weighting base classifiers
according to their prediction error rate on the latest batch of data. If the
error rate of the youngest classifier exceeds 50\%, a new classifier will be
trained based on the latest data. This method has several other benefits: it can
easily adopt almost any base classifier algorithm; it does not store history
data, only the latest batch of data, which it only uses once to train a new
classifier; and it can handle sudden drift, gradual drift, and recurrent drift
because underperforming classifiers can be reactivated or deactivated as needed
by adjusting their weights. Other voting strategies than standard weighted
voting have also been applied to handle concept drift. Examples include
hierarchical ensemble structure \cite{yin2015de2, zhang2011enabling}, short term
and long term memory \cite{Viktor:SAMkNN, xu2017concept} and dynamic ensemble
sizes \cite{pietruczuk2016a, you2016a}.

A number of research efforts have been made that focus on developing ensemble
methods for handling concept drift of certain types. \textbf{A}ccuracy
\textbf{U}pdate \textbf{E}nsemble (AUE2) \cite{Brzezinski:AUE2} was proposed
with an emphasis on handling both sudden drift and gradual drift equally well.
It is a batch mode weighted voting ensemble method based on incremental base
classifiers. By doing re-weighting, the ensemble is able react quickly to sudden
drift. All classifiers are also incrementally trained with the latest data,
which ensures that the ensemble evolves with gradual drift. The Optimal Weights
Adjustment (OWA) method \cite{zhang2008categorizing} achieves the same goal by
building ensembles using both weighted instances and weighted classifiers for
different concept drift types. The authors of \cite{sun2016online} considered a
special case of concept drift --- class evolution --- the phenomenon of class
emergence and disappearance. Recurring concepts are handled in
\cite{gama2013recurrent, gomes2014mining}, which monitor concept information to
decide when to reactivate previously stored obsolete models. \cite{Ahmadi2017}
is another method that handles recurring concepts by refining the concept pool
to avoid redundancy.

\subsection{Adjusting existing models for regional drift}
\label{sub-ka-adaptive}

\begin{figure}[!t]
\centering
\includegraphics[width=0.48\textwidth]{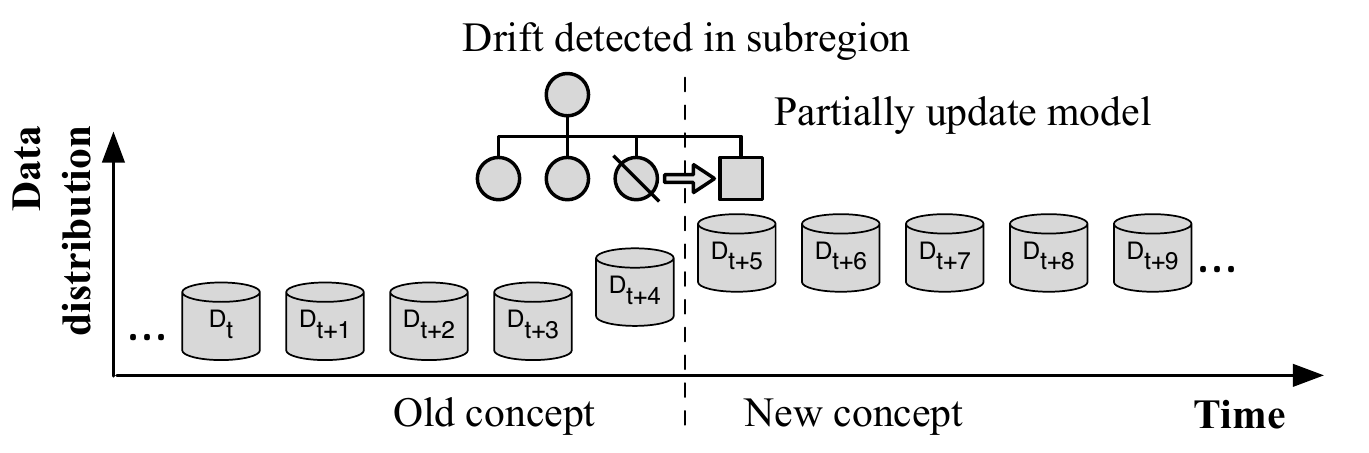}
\caption{A decision tree node is replaced with a new one as its performance
  deteriorates when a concept drift occurs in a subregion.}
\label{fig-drift-adaptation-partial}
\end{figure}

An alternative to retraining an entire model is to develop a model that
adaptively learns from the changing data. Such models have the ability to
partially update themselves when the underlying data distribution changes, as
shown in \figurename~\ref{fig-drift-adaptation-partial}. This approach is
arguably more efficient than retraining when the drift only occurs in local
regions. Many methods in this category are based on the decision tree algorithm
because trees have the ability to examine and adapt to each sub-region
separately.

In a foundational work \cite{domingos2000mining}, an online decision tree
algorithm, called \textbf{V}ery \textbf{F}ast \textbf{D}ecision \textbf{T}ree
classifier (VFDT) was proposed, which is especially tailored for high speed data
streams. It uses Hoeffding bound to limit the number of instances required for
node splitting. This method has become very popular because of its several
distinct advantages: 1) it only needs to process each instance once and does not
store instances in memory or disk; 2) the tree itself only consumes a small
amount of space and does not grow with the number of instances it processes
unless there is new information in the data; 3) the cost of tree maintenance is
very low. An extended version, called CVFDT \cite{H:tree}, was later proposed to
handle concept drift. In CVFDT, a sliding window is maintained to hold the
latest data. An alternative sub-tree is trained based on the window and its
performance is monitored. If the alternative sub-tree outperforms its original
counterpart, it will be used for future prediction and the original obsolete
sub-tree will be pruned. VFDTc \cite{gama2003accurate} is another attempt to
make improvements to VFDT with several enhancements: the ability to handle
numerical attributes; the application of naive Bayes classifiers in tree leaves
and the ability to detect and adapt to concept drift. Two node-level drift
detection methods were proposed based on monitoring differences between a node
and its sub-nodes. The first method uses classification error rate and the
second directly checks distribution difference. When a drift is detected on a
node, the node becomes a leaf and its descending sub-tree is removed. Later work
\cite{hang2012incrementally, yang2015countering} further extended VFDTc using an
adaptive leaf strategy that chooses the best classifier from three options:
majority voting, Naive Bayes and Weighted Naive Bayes.

Despite the success of VFDT, recent studies \cite{rutkowski2014decision,
  rutkowski2013decision} have shown that its foundation, the Hoeffding bound,
may not be appropriate for the node splitting calculation because the variables
it computes, the information gain, are not independent. A new online decision
tree model \cite{rutkowski2015new} was developed based on an alternative
impurity measure. The paper shows that this measure also reflects concept drift
and can be used as a replacement measure in CVFDT. In the same spirit, another
decision tree algorithm (IADEM-3) \cite{frias2016online} aims to rectify the use
of Hoeffding bound by computing the sum of independent random variables, called
\textit{relative frequencies}. The error rate of sub-trees are monitored to
detect drift and are used for tree pruning.


\section{Evaluation, Datasets and Benchmarks}
\label{sec-eb}

Section~\ref{sub-eb-evaluation} discusses the evaluation systems used for learning algorithms handling concept drift. Section~\ref{sub-eb-synthetic} introduces synthetic datasets, which used to simulate specific and controllable types of concept drift. Section~\ref{sub-eb-real-world} describes real-world datasets, which used to test the overall performance in a real-life scenario.

\subsection{Evaluation Systems}
\label{sub-eb-evaluation}


The evaluation systems is an important part for learning algorithms. Some evaluation methodologies used in learning under concept drift have been mentioned in \cite{gama2014survey}. We enrich this previous research by reviewing the evaluation systems from three aspects: 1) validation methodology, 2) evaluation metrics, and 3) statistical significance, and each evaluation is followed by its computation equation and usage introduction.

Validation methodology refers to the procedure for a learning algorithm to determine which data instances are used as the training set and which are used as the testing set. There are three procedures peculiar to the evaluation for learning algorithms capable of handling concept drift: 1) \textit{holdout}, 2) \textit{prequential}, and 3) \textit{controlled permutation}. In the scenario of a dataset involving concept drift, \textit{holdout} should follow the rule: when testing a learning algorithm at time $\T$, the holdout set represents exactly the same concept at that time $\T$. Unfortunately, it is only applied on synthetic datasets with predefined concept drift times. \textit{Prequential} is a popular evaluation scheme used in streaming data. Each data instance is first used to test the learning algorithm, and then to train the learning algorithm. This scheme has the advantage that there is no need to know the drift time of concepts, and it makes maximum use of the available data.
The prequential error is computed based on an accumulated sum of a loss function between the prediction and observed label: $S=\sum^n_{t=1} f(\hat{y}_t,y_t)$.
There are three prequential error rate estimates: a landmark window (interleaved-test-then-train), a sliding window, and a forgetting mechanism \cite{gama2012on}. \textit{Controlled permutation} \cite{zliobaite2014controlled} runs multiple test datasets in which the data order has been permutated in a controlled way to preserve the local distribution, which means that data instances that were originally close to one another in time need to remain close after a permutation. Controlled permutation reduces the risk that their prequential evaluation may produce biased results for the fixed order of data in a sequence.

Evaluation metrics for datasets involving concept drift could be selected from traditional accuracy measures, such as precision/recall in retrieval tasks, mean absolute scaled error in regression, or root mean square error in recommender systems. In addition to that, the following measures should be examined:
1) \textit{RAM-hours} \cite{bifet2010fast} for the computation cost of the mining process;
2) \textit{Kappa statistic} $\kappa = \frac{p-p_\text{ran}}{1-p_\text{ran}}$ \cite{cohen1960a} for classification taking into account class imbalance, where $p$ is the accuracy of the classifier under consideration (reference classifier) and $p_\text{ran}$ is the accuracy of the random classifier;
3) \textit{Kappa-Temporal statistic} $\kappa_{per} = \frac{p-p_\text{per}}{1-p_\text{per}}$ \cite{zliobaite2015evaluation} for the classification of streaming data with temporal dependence, where $p_\text{per}$ is the accuracy of the persistent classifier (a classifier that predicts the same label as previously observed);
4) \textit{Combined Kappa statistic} $\kappa^+ = \sqrt{\max(0,\kappa)\max(0,\kappa_\text{per})} $\cite{zliobaite2015evaluation}, which combines the $\kappa$ and $\kappa_\text{per}$ by taking the geometric average;
 5) \textit{Prequential AUC} \cite{brzezinski2014prequential}; and \hly{6) the Averaged Normalized Area Under the Curve (NAUC) values for Precision-Range curve and Recall-Range curve \cite{yu2018request}}, for the classification of streaming data involving concept drift. Apart from evaluating the performance of learning algorithms, the accuracy of the concept drift detection method/algorithm can be accessed according to the following criteria: 1) \textit{true detection rate}, 2) \textit{false detection rate}, 3) \textit{miss detection rate}, and 4) delay of detection \cite{D:kdqTree}.

Statistical significance is used to compare learning algorithms on achieved error rates. The three most frequently used statistical tests for comparing two learning algorithms \cite{bifet2015efficient, japkowicz2011evaluating} are:
1) McNemar test \cite{mcnemar1947note}: denote the number of data instances misclassified by the first classifier and correctly classified by the second classifier by $a$, and denote $b$ in the opposite way. The McNemar statistic is computed as $M=\text{sign}(a-b)\times(a-b)^2/(a+b)$ to test whether two classifiers perform equally well. The test follows the $\chi^2$ distribution;
2) Sign test: for $N$ data instances, denote the number of data instances misclassified by the first classifier and correctly classified by the second classifier by $B$ and the number of ties by $T$. Conduct one-sided sign test by computing $p=\sum_{k=B}^{N-T} \binom{N-T}{k} 0.5^k \times0.5^{N-T-k}$. If $p$ less than a significant level, then the second classifier is better than the first classifier.
and 3) Wilcoxon's sign-rank test: For testing two classifiers on $N$ datasets, let $x_{i,1}$ and $x_{i,2}$ $(i=1,\dots,N)$ denote the measurements. The number of ties is $T$ and $N_r = N - T$. The test statistic $W=\sum_{i=1}^{N_r}(\text{sign}(x_{i,1}-x_{i,2}) \times R_i)$  where $R_i$ is the rank ordered by $\left|x_{i,1} - x_{i,2}\right|$ increasingly. Two classifiers perform equally is rejected if $\left|W\right| > W_{\text{critical},N_r}$ (two-sided), where $W_{\text{critical},N_r}$ can be acquired from the statistical table.
All three tests are non-parametric.
The Nemenyi test \cite{demsar2006statistical} is used to compare more than two learning algorithms. It is an appropriate test for comparing all learning algorithms with multiple datasets, based on the average rank of algorithms over all datasets. The Nemenyi test consists of the following: two classifiers are performing differently if the corresponding average ranks differ by at least the critical
difference $\text{CD} = q_\alpha \sqrt{k(k + 1)/6N}$, where $k$ is the number of learners, $N$ is the number of datasets, and critical values $q_\alpha$ are based on the Studentized range statistic divided by $\sqrt{2}$.
Other tests can be used to compare learning algorithms with a control algorithm \cite{demsar2006statistical}.

\subsection{Synthetic datasets}
\label{sub-eb-synthetic}

We list several widely used synthetic datasets for evaluating the performance of learning algorithms dealing with concept drift. Since data instances are generated by predefined rules and specific parameters, a synthetic dataset is a good option for evaluating the performance of learning algorithms in different concept drift scenarios. The dataset provider, the number of instances (\#Insts.), the number of attributes (\#Attrs.), the number of classes (\#Cls.), types of drift (Types), sources of drift (Sources), and used by references, are listed in \tablename~\ref{tab-synthetic}.

\begin{table*}
\renewcommand{\arraystretch}{1.1}
\caption{List of synthetic datasets for performance evaluation of learning under concept drift.}
\label{tab-synthetic}
\centering
\scriptsize
\begin{tabular}{|p{0.2cm}|p{3cm}|p{0.7cm}|p{0.7cm}|p{0.7cm}|p{3cm}|p{0.7cm}|p{5.5cm}|}
\hline    & Dataset                                         & \#Insts. & \#Attrs. & \#Cls. & Types                        & Sources                       & Used by references \\
\hline 1  & STAGGER\cite{W:machinelearning}                 & Custom   & 3        & 2      & Sudden                       & II          & \cite{bach2008paired, Alippi:LSDD-INC, frias2016online, I:HDDM, Gama:DDM, kolter2004using, li2015learning, N:STEPD, XD:SUN, Xu:DELM, you2016a} \\
\hline 2  & SEA\cite{S:SEA}                                 & Custom   & 3        & 2      & Sudden                       & II          & \cite{bach2008paired, Bifet:ADWIN_HWT, bifet2010leveragin, Brzezinski:AUE2, elwell2011incremental, fok2017mining, gama2013recurrent, Gama:DDM, Bifet:ADWIN_ARF, Kosina2015, li2015learning, Liu:FuzzyWindow, liu2016fpelm, Viktor:SAMkNN, L:AI2, S:SEA, XD:SUN, Xu:DELM, Yu:HLFR} \\
\hline 3  & Rotating hyperplane\cite{H:tree, ensembleWCE}   & Custom   & 10       & 2      & Gradual; Incremental         & II          & \cite{Ahmadi2017, Bifet:ADWIN_HWT, bifet2010leveragin, Brzezinski:AUE2, Alippi:LSDD-CDT, Alippi:LSDD-INC, frias2016online, Bifet:ADWIN_ARF, Han2015, H:tree, Kosina2015, liu2016fpelm, Viktor:SAMkNN, L:AI2, N:STEPD, pietruczuk2016a, J:PCAdriftDetection, Xu:DELM, yang2015countering, you2016a, Yu:HLFR} \\
\hline 4  & Random RBF\cite{B:MOA}                          & Custom   & Custom   & Custom & Sudden; Gradual; Incremental & III         & \cite{Alippi:HCDTs, B:EDDM, bifet2010leveragin, Brzezinski:AUE2, Alippi:LSDD-CDT, Alippi:LSDD-INC, elwell2011incremental, frias2016online, Gama:DDM, Bifet:ADWIN_ARF, Kosina2015, Viktor:SAMkNN, N:STEPD, G:ECDD, souza2015data, Xu:DELM, you2016a, zhang2008categorizing} \\
\hline 5  & Random Tree\cite{B:MOA, domingos2000mining}     & Custom   & Custom   & Custom & Sudden; Reoccurring          & II          & \cite{Brzezinski:AUE2, frias2016online, Bifet:ADWIN_ARF, rutkowski2015new, rutkowski2014decision, rutkowski2013decision, Xu:DELM, hang2012incrementally} \\
\hline 6  & LED\cite{UCIDataset}                            & Custom   & 24       & 10     & Sudden                       & II          & \cite{bifet2010leveragin, Brzezinski:AUE2, frias2016online, I:HDDM, gama2003accurate, Bifet:ADWIN_ARF, Kosina2015, XD:SUN, Xu:DELM, hang2012incrementally} \\
\hline 7  & Waveform\cite{UCIDataset}                       & Custom   & 40       & 3      & Sudden                       & II          & \cite{Ahmadi2017, Dries:MMDconceptDriftDetection, frias2016online, gama2003accurate, Kosina2015, Xu:DELM, hang2012incrementally, yang2015countering} \\
\hline 8  & Sine\cite{Gama:DDM}                             & Custom   & 2        & 2      & Sudden                       & II          & \cite{B:EDDM, Alippi:LSDD-CDT, Gama:DDM, Harel:Gradual, G:ECDD, you2016a} \\
\hline 9  & Circle\cite{Gama:DDM}                           & Custom   & 2        & 2      & Gradual                      & III         & \cite{B:EDDM, Alippi:LSDD-CDT, Alippi:LSDD-INC, fok2017mining, Gama:DDM, Harel:Gradual, N:STEPD, you2016a} \\
\hline 10 & Rotating chessboard\cite{elwell2011incremental} & Custom   & 2        & 2      & Gradual                      & II          & \cite{Alippi:JustInTimeRecurrentDrift, elwell2011incremental, Harel:Gradual, Viktor:SAMkNN, Yu:HLFR} \\
\hline
\end{tabular}

\end{table*}

\subsection{Real-world datasets}
\label{sub-eb-real-world}

In this section, we collect several publicly available real-world datasets, including real-world datasets with synthetic drifts. The dataset provider, the number of instances (\#Insts.), the number of attributes (\#Attrs.), the number of classes (\#Cls.), and used by references, are shown in \tablename~\ref{tab-real-world}.

\hly{Most of these datasets contain temporal concept drift spanning over different period range - e.g. daily (Sensor \mbox{\cite{ZhuDataset}}), seasonally (Electricity \cite{harries1999splice2}) or yearly (Airlines \mbox{\cite{B:MOA}}, NOAA weather\cite{elwell2011incremental}). Others include geographical (Covertype\cite{UCIDataset}) or categorical (Poker-Hand\cite{UCIDataset}) concept drift. Certain datesets, mainly text based, are targeting at specific drift types, such as sudden drift (Email\_data\cite{Katakis2010}), gradural drift (Spam assassin corpus\cite{K:Spam500}), recurrent drift (Usenet \mbox{\cite{IK:Usenet}}) or novel class (KDDCup'99\cite{UCIDataset}, ECUE drift dataset 2\cite{D:firstSpam})

These datasets provide realistic benchmark for evaluating differnent concept drift handling methods. There are, however, two limitations of real world data sets: 1) the groud truth of precise start and end time of drifts is unknown; 2) some real datasets may include mixed drift types. These limitations make it difficult to evaluate methods for understanding the drift, and could introduce bias when comparing different machine learning models.}

\begin{table*}
\renewcommand{\arraystretch}{1.1}
\caption{List of real-world datasets for performance evaluation of learning under concept drift.}
\label{tab-real-world}
\centering
\scriptsize
\begin{tabular}{|p{0.2cm}|p{3.2cm}|p{1cm}|p{1cm}|p{0.5cm}|p{9.5cm}|}
\hline    & Dataset                                  & \#Insts. & \#Attrs.  & \#Cls.    & Used by references \\
\hline 1  & Airlines\cite{B:MOA}                     & 539384   & 7         & 2         & \cite{Brzezinski:AUE2, Bifet:ADWIN_ARF, Kosina2015, Liu:IJCAI2017, Liu:FuzzyWindow, wang2017tracking, Z:ActiveLearning} \\
\hline 2  & Covertype\cite{UCIDataset}               & 581012   & 54        & 7         & \cite{bifet2010leveragin, Brzezinski:AUE2, frias2016online, I:HDDM, gama2003accurate, Bifet:ADWIN_ARF, Han2015, Kosina2015, Viktor:SAMkNN, rutkowski2015new, J:PCAdriftDetection, hang2012incrementally, yang2015countering, Z:ActiveLearning} \\
\hline 3  & Electricity \cite{harries1999splice2}    & 45312    & 8         & 2         & \cite{Ahmadi2017, bach2008paired, B:EDDM, Bifet:ADWIN, bifet2010leveragin, Brzezinski:AUE2, fok2017mining, frias2016online, I:HDDM, Gama:DDM, Bifet:ADWIN_ARF, Kosina2015, Liu:IJCAI2017, Liu:FuzzyWindow, Viktor:SAMkNN, G:ECDD, rutkowski2015new, J:PCAdriftDetection, you2016a, Z:ActiveLearning} \\
\hline 4  & Poker-Hand\cite{UCIDataset}              & 1025010  & 10        & 10        & \cite{Bifet:ADWIN_HWT, bifet2010leveragin, Brzezinski:AUE2, Kosina2015, Viktor:SAMkNN} \\
\hline 5  & NOAA weather\cite{elwell2011incremental} & 18159    & 8         & 2         & \cite{Ahmadi2017, elwell2011incremental, Liu:IJCAI2017, Viktor:SAMkNN, L:AI2, souza2015data, yin2015de2} \\
\hline 6  & Sensor\cite{ZhuDataset}                  & 2219803  & 5         & 54        & \cite{Ahmadi2017, J:PCAdriftDetection} \\
\hline 7  & KDDCup'99\cite{UCIDataset}               & 494021   & 41        & 23        & \cite{Bifet:ADWIN_ARF, Kosina2015, li2015learning, rutkowski2015new, rutkowski2014decision, XD:SUN, zhang2011enabling, zhang2008categorizing, PeiPei:ThreeLayer} \\
\hline 8  & Usenet1\cite{IK:Usenet}                  & 1500     & 99        & 2         & \cite{frias2016online, I:HDDM, Yu:HLFR} \\
\hline 9  & Usenet2\cite{IK:Usenet}                  & 1500     & 99        & 2         & \cite{frias2016online, I:HDDM} \\
\hline 10 & Email\_data\cite{Katakis2010}            & 1500     & 913       & 2         & \cite{Alippi:JustInTimeRecurrentDrift, gama2013recurrent, gomes2014mining} \\
\hline 11 & Spam\_data\cite{Katakis2010}             & 9324     & 499       & 2         & \cite{I:HDDM, Kosina2015, Liu:IJCAI2017, Liu:FuzzyWindow, J:PCAdriftDetection, song2016dynamic} \\
\hline 12 & Spam assassin corpus\cite{K:Spam500}     & 9324     & 39916     & 2         & \cite{frias2016online, gama2013recurrent, Bifet:ADWIN_ARF, Liu:IJCAI2017} \\
\hline 13 & ECUE drift dataset 1\cite{D:firstSpam}   & 10983    & 287034    & 2         & \cite{L:AI2, L:competence} \\
\hline 14 & ECUE drift dataset 2\cite{D:firstSpam}   & 11905    & 166047    & 2         & \cite{L:AI2, L:competence} \\
\hline
\end{tabular}

\end{table*}

\section{The Concept Drift Problem in Other Research Areas}
\label{sec-ra}

We have observed that handling the concept drift problem is not a standalone research subject but has a large number of indirect usage scenarios. In this section, we adopt this new perspective to review recent developments in other research areas that benefit from handling the concept drift problem.

\subsection{Class imbalance}
\label{sub-ra-class-imbalance}

Class imbalance is a common problem in stream data mining in addition to concept drift. Research effort has been made to develop effective learning algorithms to tackle both problems at same time. \cite{ditzler2013incremental} presented two ensemble methods for learning under concept drift with imbalanced class. The first method, Learn++.CDS, is extended from Learn++.NSE and combined with the Synthetic Minority class Oversampling Technique (SMOTE). The second algorithm, Learn++.NIE, improves on the previous method by employing a different penalty constraint to prevent prediction accuracy bias and replacing SMOTE with bagging to avoid oversampling. ESOS-ELM \cite{mirza2015ensemble} is another ensemble method which uses Online Sequential Extreme Learning Machine (OS-ELM) as a basic classifier to improve performance with class imbalanced data. A concept drift detector is integrated to retrain the classifier when drift occurs. The author then developed another algorithm \cite{mirza2016meta}, which is able to tackle multi-class imbalanced data with concept drift. \cite{wang2015resampling} proposed two learning algorithms OOB and UOB, which build an ensemble model to overcome the class imbalance in real time through resampling and time-decayed metrics. \cite{arabmakki2017som} developed an ensemble method which handles concept drift and class imbalance with additional true label data limitation.

\subsection{Big data mining}
\label{sub-ra-bigdata}

Data mining in big data environments faces similar challenges to stream data mining \cite{katal2013big}: data is generated at a fast rate (Velocity) and distribution uncertainty always exists in the data, which means that handling concept drift is also crucial in big data applications. Additionally, scalability is an important consideration because in big data environments, a data stream may come in very large and potentially unpredictable quantities (Volume) and cannot be processed in a single computer server. An attempt to handle concept drift in a distributed computing environment was made by \cite{andrzejak2012parallel} in which an Online Map-Reduce Drift Detection Method (OMR-DDM) was proposed, using the combined online error rate of the parallel classification algorithms to identify the changes in a big data stream. A recent study \cite{tennant2017scalable} proposed another scalable stream data mining algorithm, called Micro-Cluster Nearest Neighbor (MC-NN), based on nearest neighbor classifier. This method extends the original Micro-Cluster algorithm \cite{A:CluStream} to adapt to concept drift by monitoring classification error. This micro-cluster algorithm was further extended to a parallel version using the map-reduce technique in \cite{song2016a} and applied to solve the label-drift classification problem where class labels are not known in advance \cite{nguyen2016one}.

\subsection{Active learning and semi-supervised learning}
\label{sub-ra-active}

Active learning is based on the assumption that there is a large amount of unlabeled data but only a fraction of them can be labeled by human effort. This is a common situation in stream data applications, which are often also subject to the concept drift problem. \cite{Z:ActiveLearning} presented a general framework that combines active learning and concept drift adaptation. It first compares different instance-sampling strategies for labeling to guarantee that the labeling cost will be under budget, and that distribution bias will be prevented. A drift adaptation mechanism is then adopted, based on the DDM detection method \cite{Gama:DDM}. In \cite{chu2011unbiased}, the authors proposed a new active learning algorithm that primarily aims to avoid bias in the sampling process of choosing instances for labeling. They also introduced a memory loss factor to the model, enabling it to adapt to concept drift.

Semi-supervised learning concerns how to use limited true label data more efficiently by leveraging unsupervised techniques. In this scenario, additional design effort is required to handle concept drift. For example, in \cite{ditzler2011semi}, the authors applied a Gaussian Mixture model to both labeled and unlabeled data, and assigned labels, which has the ability to adapt to gradual drift. Similarly, \cite{Hosseini2016, XD:SUN, zhang2010classifier} are all cluster-based semi-supervised ensemble methods that aim to adapt to drift with limited true label data. The latter are also able to recognize recurring concepts. In \cite{chandra2016adaptive}, the author adopted a new perspective on the true label scarcity problem by considering the true labeled data and unlabeled data as two independent non-stationary data generating processes. Concept drift is handled asynchronously on these two streams. The SAND algorithm \cite{haque2016efficient, haque2016sand} is another semi-supervised adaptive method which detects concept drift on cluster boundaries. There are also studies [90, 91] that focus on adapting to concept drift in circumstances where true label data is completely unavailable.

\subsection{Decision Rules}
\label{sub-ra-rule}

Data-driven decision support systems need to be able to adapt to concept drift in order to make accurate decisions and decision rules is the main technique for this purpose. \cite{Kosina2015} proposed a decision rule induction algorithm, Very Fast Decision Rules (VFDR), to effectively process stream data. An extended version, Adaptive VFDR, was developed to handle concept drift by dynamically adding and removing decision rules according to their error rate which is monitored by drift detector. Instead of inducing rules from decision trees, \cite{le2017on} proposed another decision rule algorithm based on PRISM \cite{cendrowska1987prism} to directly induce rules from data. This algorithm is also able to adapt to drift by monitoring the performance of each rule on a sliding window of latest data. \cite{pratama2015pclass} also developed an adaptive decision making algorithm based on fuzzy rules. The algorithm includes a rule pruning procedure, which removes obsolete rules to adapt to changes, and a rule recal procedure to adapt to recurring concepts.

This section by no means attempts to cover every research field in which concept drift handling is used. There are many other studies that also consider concept drift as a dual problem. For example, \cite{yeh2013rank} is a dimension reduction algorithm to separate classes based on least squares linear discovery analysis (LSLDA), which is then extended to adapt to drift; \cite{cavalcante2016fedd} considered the concept drift problem in time series and developed an online explicit drift detection method by monitoring time series features; and \cite{pratama2016scaffolding} developed an incremental scaffolding classification algorithm for complex tasks that also involve concept drift.

\section{Conclusions: findings and future directions}
\label{sec-ca}

We summarize the recent developments of concept drift research, and the following important findings can be extracted: 

\begin{enumerate}[\IEEEsetlabelwidth{5)}]

\item Error rate-based and data distribution-based drift detection methods are still playing a dominant role in concept drift detection research, while multiple hypothesis test methods emerge in recent years; 

\item Regarding to concept drift understanding, all drift detection methods can answer ``When'', but very few methods have the ability to answer ``How'' and ``Where'';

\item Adaptive models and ensemble techniques have played an increasingly important role in recent concept drift adaptation developments. In contrast, research of retraining models with explicit drift detection has slowed;

\item \hly{Most existing drift detection and adaptation algorithms assume the ground true label is available after classification/prediction, or extreme verification latency. Very few research has been conducted to address unsupervised or semi-supervised drift detection and adaptation.}

\item Some computational intelligence techniques, such as fuzzy logic, competence model, have been applied in concept drift;

\item There is no comprehensive analysis on real-world data streams from the concept drift aspect, such as the drift occurrence time, the severity of drift, and the drift regions.

\item An increasing number of other research areas have recognized the importance of handling concept drift, especially in big data community.

\end{enumerate}


Based on these findings, we suggest four new directions in future concept drift research:

\begin{enumerate}[\IEEEsetlabelwidth{3)}]

\item Drift detection research should not only focus on identifying drift occurrence time accurately, but also need to provide the information of drift severity and regions. These information could be utilized for better concept drift adaptation.

\item \hly{In the real-world scenario, the cost to acquire true label could be expensive, that is, unsupervised or semi-supervised drift detection and adaptation could still be promising in the future.}

\item A framework for selecting real-world data streams should be established for evaluating learning algorithms handling concept drift.

\item Research on effectively integrating concept drift handling techniques with machine learning methodologies for data-driven applications is highly desired.

\end{enumerate}

We hope this paper can provide researchers with state-of-the-art knowledge on concept drift research developments and provide guidelines about how to apply concept drift techniques in different domains to support users in various prediction and decision activities. 
\section*{Acknowledgments}

The work presented in this paper was supported by the Australian Research Council (ARC) under discovery grant DP150101645. We significantly thank Yiliao Song for her help in preparation of datasets and applications shown in Sections \ref{sec-eb}.

\ifCLASSOPTIONcaptionsoff
  \newpage
\fi



\bibliographystyle{./bib/IEEEtran}
\bibliography{IEEEabrv,survey,addRef}
%



%


\begin{IEEEbiography}[{\includegraphics[width=1in,height=1.25in,clip,keepaspectratio]{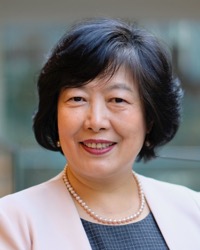}}]{Jie Lu}
is a Distinguished Professor, Director of Centre for Artificial Intelligence, and Associate Dean Research with in the Faculty of Engineering and Information Technology at the University of Technology Sydney. Her research interests lie in the area of decision support systems, concept drift, fuzzy transfer learning, and recommender systems. She has published 10 research books and 400 papers, won 8 Australian Research Council discovery grants and 20 other grants. She serves as Editor-In-Chief for KBS and IJCIS, and delivered 16 keynotes in international conferences. \vspace{-15 mm}
\end{IEEEbiography}

\begin{IEEEbiography}[{\includegraphics[width=1in,height=1.25in, clip]{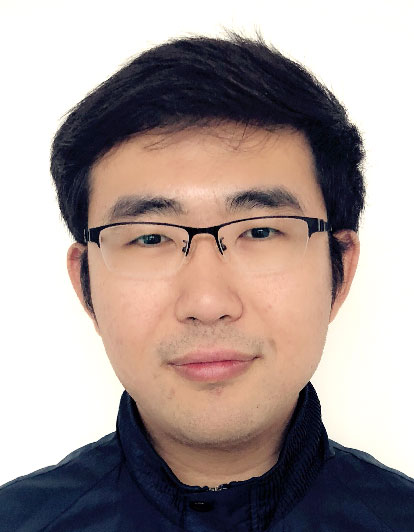}}]{Anjin Liu} is a Postdoctoral Research Associate in the A/DRsch Centre for Artificial Intelligence, Faculty of Engineering and Information Technology, University of Technology Sydney. He received the BIT degree (Honour) at the University of Sydney in 2012. His research interests include concept drift detection, adaptive data stream learning, multi-stream learning, machine learning and big data analytics \vspace{-15 mm}
\end{IEEEbiography}

\begin{IEEEbiography}[{\includegraphics[width=1in,height=1.25in, clip]{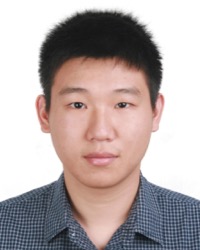}}]{Fan Dong} is Research Fellow of Centre for Artificial Intelligence, University of Technology Sydney. He received the dual Ph.D. degree from University of Technology Sydney and Beijing Institute of Technology in 2018. His research interests include concept drift detection, adaptive learning under concept drift and data stream mining. \vspace{-15 mm}
\end{IEEEbiography}

\begin{IEEEbiography}[{\includegraphics[width=1in,height=1.25in, clip]{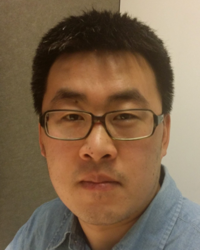}}]{Feng Gu} is a Ph.D. candidate at the Faculty of Engineering and Information Technology, the University of Technology Sydney, NSW, Australia. He received bachelor’s degree of software engineering at Zhejiang University, China, in 2012. His research interests include stream data mining, adaptive learning under concept drift and evolving data. \vspace{-15 mm}
\end{IEEEbiography}

\begin{IEEEbiography}[{\includegraphics[width=1in,height=1.25in, clip]{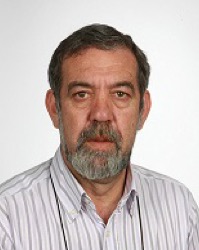}}]{Jo\~{a}o~Gama} is an Associate Professor at the University of Porto, Portugal. He is also a senior researcher and member of the board of directors of the Laboratory of Artificial Intelligence and Decision Support (LIAAD), a group belonging to INESC Porto. He serves as the member of the Editorial Board of Machine Learning Journal, Data Mining and Knowledge Discovery, Intelligent Data Analysis and New Generation Computing.
His main research interest is in knowledge discovery from data streams and evolving data. He has published more than 200 papers and a recent book on Knowledge Discovery from Data Streams. He has extensive publications in the area of data stream learning. \vspace{-10 mm}
\end{IEEEbiography}

\begin{IEEEbiography}[{\includegraphics[width=1in,height=1.25in, clip]{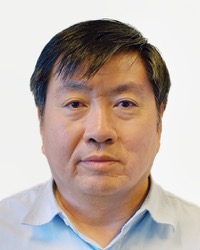}}]{Guangquan Zhang}
is an Associate Professor, and the Director of Decision System and e-Service Intelligence (DeSI) lab with in the Centre for Artificial Intelligence, in the Faculty of Engineering and Information Technology at the University of Technology Sydney. His main research interests lie in the area of uncertain information processing, fuzzy decision making, concept drift and fuzzy transfer learning. He has published 4 monographs and over 400 papers in refereed journals, conference proceedings and book chapters. He has won 7 Australian Research Council discovery grants and guest edited many special issues for international journals. \vspace{-10 mm}
\end{IEEEbiography}



\end{document}